\documentclass{article} 
\usepackage{epsfig,amsfonts,color}
\usepackage{amsmath}
\usepackage{amssymb, palatino, geometry,url}
\usepackage{multirow}
\usepackage{lscape}

\usepackage{graphicx}
\usepackage[margin = 1cm]{caption}
\usepackage{subcaption}
\usepackage{placeins}
\usepackage[colorlinks=true,linkcolor=blue,citecolor=blue,urlcolor=blue]{hyperref}
\usepackage[titletoc,title]{appendix}
\usepackage{pdflscape}

\usepackage{listings}
\DeclareCaptionFont{white}{\color{white}}
\DeclareCaptionFormat{listing}{\colorbox[cmyk]{0.7, 0.35, 0.35,0.01}{\parbox{\dimexpr\textwidth-2\fboxsep\relax}{#1#2#3}}}

\usepackage[colorlinks=true,linkcolor=blue,citecolor=blue,urlcolor=blue]{hyperref}

\usepackage{fancyhdr}
\usepackage{lineno}

\begin{document}

\title{A Graph-Based Modeling Framework for Tracing \\ Hydrological Pollutant Transport in Surface Waters}

\author{David L. Cole${}^{a}$, Gerardo J. Ruiz-Mercado${}^{bc}$, and Victor M. Zavala${}^{a}$\thanks{Corresponding Author:\\ victor.zavala@wisc.edu \\ 1415 Engineering Drive \\ Madison, WI 53706}
}

\date{\small
  ${}^a$Department of Chemical and Biological Engineering, \\[0in]
  University of Wisconsin-Madison, Madison, WI 53706 \\
  ${}^b$Office of Research and Development, \\
  U.S. Environmental Protection Agency, Cincinnati, OH 45268, USA,\\
  and Chemical Engineering Graduate Program, \\
  Universidad del Atl\'antico, Puerto Colombia 080007, Colombia
}

\maketitle

\begin{abstract}
Anthropogenic pollution of hydrological systems affects diverse communities and ecosystems around the world. Data analytics and modeling tools play a key role in fighting this challenge, as they can help identify key sources as well as trace transport and quantify impact within complex hydrological systems. Several tools exist for simulating and tracing pollutant transport throughout surface waters using detailed physical models; these tools are powerful, but can be computationally intensive, require significant amounts of data to be developed, and require expert knowledge for their use (ultimately limiting application scope). In this work, we present a graph modeling framework---which we call {\tt HydroGraphs}---for understanding pollutant transport and fate across waterbodies, rivers, and watersheds. This framework uses a simplified representation of hydrological systems that can be constructed based purely on open-source data (National Hydrography Dataset and Watershed Boundary Dataset). The graph representation provides a flexible intuitive approach for capturing connectivity and for identifying upstream pollutant sources and for tracing downstream impacts within small and large hydrological systems. Moreover, the graph representation can facilitate the use of advanced algorithms and tools of graph theory, topology, optimization, and machine learning to aid data analytics and decision-making. We demonstrate the capabilities of our framework by using case studies in the State of Wisconsin; here, we aim to identify upstream nutrient pollutant sources that arise from agricultural practices and trace downstream impacts to waterbodies, rivers, and streams. Our tool ultimately seeks to help stakeholders design effective pollution prevention/mitigation practices and evaluate how surface waters respond to such practices. 
\end{abstract}

\\

\section*{Highlights}
\begin{itemize}
    \item We present a general framework for representing hydrological systems as graphs
    \item The graph-based framework can help identify pollutant transport and fate across waterbodies, rivers, and watersheds
    \item Representing these systems as graphs enables advanced metrics and algorithms to aid data analytics and decision making
    \item We apply our framework to case studies in Wisconsin, USA to highlight applications
\end{itemize}

\section{Introduction}

Anthropogenic pollution in hydrological systems has significant impacts on communities and ecosystems around the globe. These pollutants arise from diverse sources and come in many forms; common pollutants include  nutrients such as nitrogen- and phosphorus-based fertilizers \cite{carpenter1998nonpoint}, emerging contaminants (ECs)  \cite{wilkinson2017occurrence}, microplastics \cite{haddout2022microplastics}, heavy metals \cite{ciazela2018tracking}, and microbes \cite{nawab2016health}. These contaminants can move through surface waters (e.g., lakes, rivers, streams) and groundwaters by following complex network pathways, leading to impacts near the contaminant release as well as to downstream impacts that span thousands of miles. Studies have shown that many of these pollutants can be toxic to humans and animals, and many of these impacts are still not fully understood. For instance, ECs include chemicals such as pharmaceuticals, personal care products, and per- and poly-fluoroalkyl substances (PFAS) \cite{tong2022source}. Risk assessment studies on some ECs suggest that they can impact the immune system or cause cancer, while other ECs have little published health information \cite{bonato2020pfas,cousins2020strategies}. Other pollutants, such as heavy metals or microbes, can likewise lead to cancer or waterborne diseases, such as dysentery or diarrhea \cite{khan2013drinking, lim2008heavy}.\\
\\

Pollutants can also cause significant environmental and economic problems; for instance, nutrient pollution is a major driver of harmful algae blooms (HABs) in both freshwater and saline waterbodies \cite{nie2018causes,shortle2017nutrient, board2000clean}. Nutrient pollution and HABs can destroy marine wildlife through anoxia, poisoning, and other mechanisms \cite{bauman2010environment,brusle1994fish, rabotyagov2020deadzone}, can cause human health impact, and they can decrease property values and hurt recreational and fishing operations on the order of billions of US dollars \cite{dodds2009eutrophication, sampat2021valuing}. Many pollutants---such as ECs, microplastics, and heavy metals---have also been shown to bioaccumulate in wildlife, leading to long-term effects \cite{copat2012heavy,  zhang2019interactive, wilkinson2017occurrence}. Furthermore, the fate of some pollutants such as plastics is not fully understood; many of these contaminants have uncertain environmental and human health impacts and can degrade into smaller compounds with unknown properties \cite{gogoi2018occurrence, wilkinson2017occurrence}. These pollutants also tend to travel (via hydrological systems) over long distances and find their way to oceans \cite{ho2019widespread, rabotyagov2020deadzone}. Moreover, such pollutants may disproportionately impact vulnerable communities, such as those in rural areas and developing countries \cite{ashbolt2004microbial}.
\\

To better understand and combat the environmental, economic, and health impacts of hydrological pollutants, there is a need for models that provide intuitive and easy-to-use tools that can help navigate complexity and answer questions of interest. For instance, for a given pollutant release, it is important to understand which parts of a hydrological system will likely be impacted or how far a pollutant can travel. Similarly, if a contaminant is discovered in a given river or lake, we might be interested in identifying what are possible upstream sources from which it originated. These types of questions are challenging to address because hydrological systems involve large and highly interconnected networks. For instance, the conterminous United States contains more than 85,000 lakes \cite{king2021graph} and over one million kilometers of rivers and streams \cite{epa2020}. Moreover, interconnections between waterbodies span multiple spatial scales and are often non-intuitive. For instance, the Mississippi river is connected to waterbodies in Pennsylvania but not to waterbodies in Michigan. As a result, pollutant transport can also be complex, as many pollutants can originate from point and non-point sources \cite{carpenter1998nonpoint, nie2018causes, xue2022review}, can come from far upstream \cite{saul2019downstream}, can involve significant spatial/temporal scales (i.e., legacy pollutants) \cite{motew2017influence,li2018legacy, sharpley2013legacy}, and can be dependent on many factors such as weather, topology, soil type, or land cover \cite{sharpley1993phosphorus, van2004effect, zhu2021impact}. 
\\

Diverse tools exist for understanding pollutant transport using detailed physical models \cite{costa2021model, lindim2016large, mispan2015model,tong2022source, wellen2015model,yuan2020review}; these tools are powerful, but are computationally intensive and may require significant expertise and data to be used (ultimately limiting application scope). \cite{alam2021modelling, costa2021model, mispan2015model, tong2022source,wellen2015model,yuan2020review} each presented reviews (collectively covering dozens of models) of many pollution models which are often focused on capturing the dynamics of pollution flux throughout a given basin. These models have an important role in understanding the pollutant transport, but they are also computationally intensive as they typically include different pollutant pools (e.g., pollutants in soil, pollutants in waterbodies, etc.) with many equations to capture the spatial and temporal behavior of the pollutants (e.g., SWAT \cite{arnold1998large} or STREAM-EU \cite{lindim2016large}). Many models also require significant data about a given watershed (e.g., AGNPS \cite{young1989agnps}) or use empirical data to regress the model (e.g., SPARROW \cite{schwarz2006sparrow}). Further, because of their complexity, they often require notable expertise and/or are confined to specific study areas (i.e., it may be intractable to apply some of them to very large study areas). Consequently, we are interested in scalable tools that can more easily map pollutant sources and impacts.
\\

A modeling approach that can help trace pollutant pathways in a more simplified manner consists of representing hydrological systems as graphs (networks). Graphs are mathematical representations (models) that are comprised of sets of nodes and edges; nodes represent different objects (e.g., lakes), and edges are links placed between nodes (e.g., rivers and streams connecting lakes). A wide range of applications of graph representations have been explored in science and engineering (from cosmology \cite{krioukov2012network, villanueva2022learning} to social networks \cite{majeed2020graph} and infrastructure/ecology networks \cite{fortin2012spatial,ort2009model, wagner2010assessing,agarwal2022evaluation, luo2018explaining}). The success of graph modeling tools in such applications has been due to the availability of diverse analysis/visualization tools and of underlying algorithms that enable scalable analysis. Many tools and algorithms are specific to graph representations (e.g., spectral properties, topological data analysis, or graph convolutional neural networks), so representing hydrological systems as graphs provides access to methods that may not be available under other existing models. In addition, graphs provide an intuitive and flexible approach for analyzing connectivity, which can be an important factor in understanding pollutant impacts and transport within hydrological systems \cite{carpenter2014connectivity, cheruvelil2022, soranno2015connectivity}. 
\\

Graph models have been used in different studies to model hydrological systems, and there are some existing tools or datasets that involve graph-based models in hydrological systems. These are outlined below. 
\begin{itemize}
    \item Recently, King and co-workers \cite{king2021graph} used graphs for analyzing connectivity of lakes in the US; however, this work did not target hydrological pollutant tracing and the graph representation used did not capture rivers and streams as nodes within the graph. Other works have used graphs to represent river systems, but these did not include explicit connectivity to waterbodies or analyzed pollutant transport to waterbodies \cite{abed2017emergent, heckmann2015graph,schmidt2020microplastic, tejedor2015delta1, tejedor2015delta2, zaliapin2010transport}.
    \item There are other datasets that capture connectivity of some hydrological systems, such as the Watershed Index Online (WSIO; captures connectivity of HUC12 watersheds) \cite{WSIO2022} or the Wisconsin Department of Natural Resources (DNR) 24K Hydrography Geodatabase \cite{epa2014, WDNR24Kgeodatabase}. These sources do not explicitly represent the waterbody-river system as a graph, but they provide important data that can be incorporated into a graph model of rivers and waterbodies. As will be shown later, a graph model can provide an easy way to incorporate pollutant data, and it can be applied to areas not covered by WSIO or the Wisconsin DNR 24K Hydrography Geodatabase.
    \item ArcGIS has capabilities for working with networks which capture connectivity (e.g., their Network Dataset toolset and Network Analyst Extension). However, ArcGIS is not open-source, and their network representations are not graphs (they capture connectivity, but are not presented under a graph theory framework) which restricts some analysis techniques. To our understanding, these tools are also implemented within the US Environmental Protection Agency's (EPA) WATERS GeoViewer \cite{watersgioviewer}. This Geoviewer does capture connectivity of streams and basins and provides a great user interface. However, to our knowledge, there is not a way (without exporting to ArcGIS) to perform analysis quickly over many different locations, as can be done with a graph framework with access to graph-analysis tools.
    \item Lastly, \cite{tian2022evaluation} represented a hydrologic system in a single basin (including streams) as a graph to help identify hydrologic connectivity projects based on graph metrics. Their work was not a general framework for representing hydrologic systems as graphs, but in using graphs for their analysis, they do highlight some of the practicality of graph networks. 
\end{itemize}

\begin{figure}[!htp]
    \centering
    \includegraphics[scale=.4]{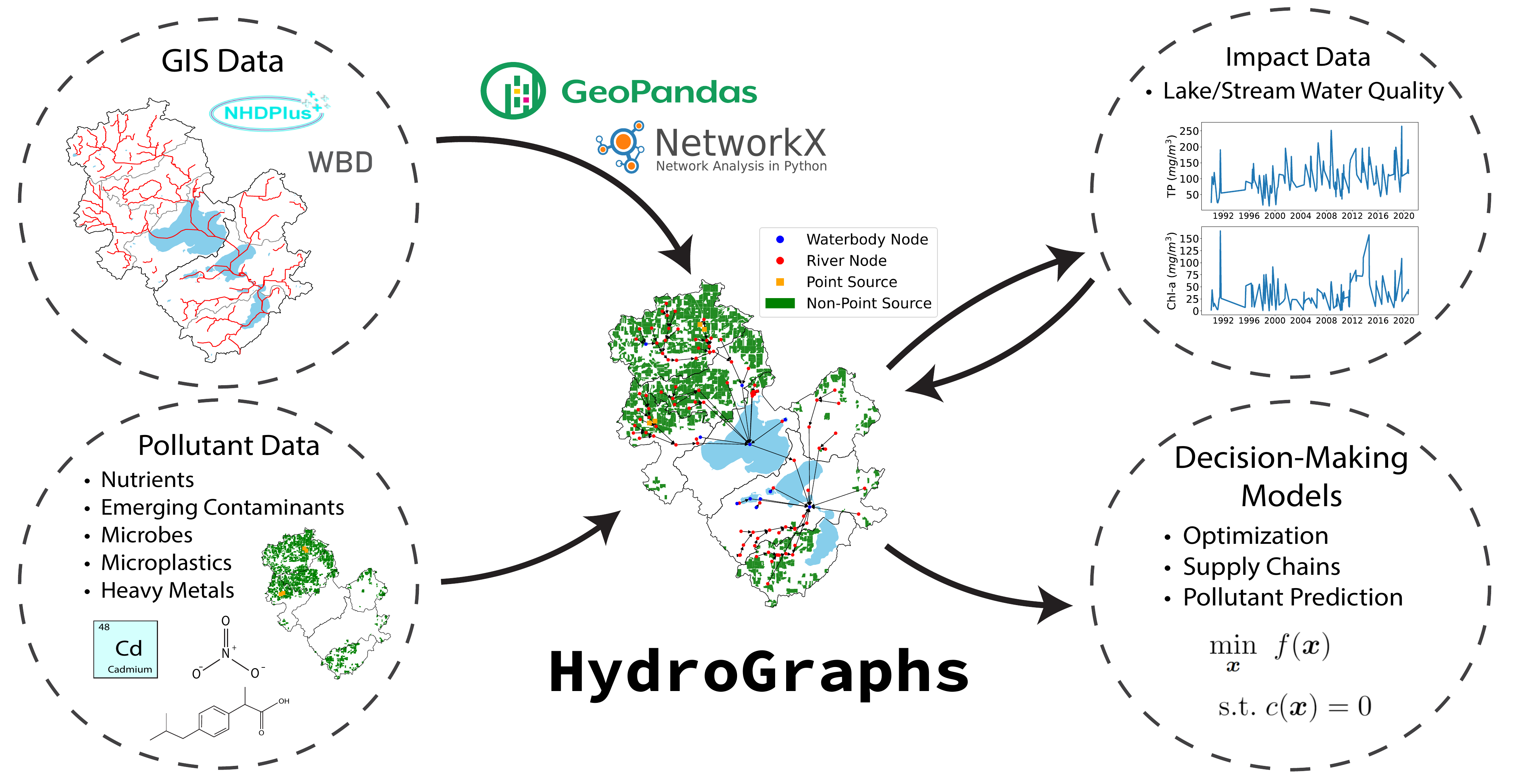}
    \caption{\footnotesize An overview of {\tt HydroGraphs}, a graph-based modeling framework for incorporating geospatial data and pollutant data to trace anthropogenic pollution transport through surface waters. The framework uses open data from the National Hydrography Dataset (NHDPlusV2) \cite{NHDPlus2019} and the Watershed Boundary Dataset \cite{usgs2019WBD}, and uses {\tt GeoPandas} \cite{geopandas} and {\tt NetworkX} \cite{NetworkX2008} for building and analyzing the resulting graph representation. Total phosphorus and chlorophyll-a data shown in the top right is from the Wisconsin Department of Natural Resources for Lake Winnebago, WI\cite{dnr}.}
    \label{fig:graphical_abstract}
  \end{figure}

In this work, we present a graph modeling framework---which we call {\tt HydroGraphs}---for capturing watershed-river-waterbody connectivity in hydrological systems. The proposed framework was implemented in Python and was developed with the goal of analyzing pollutant transport (Figure \ref{fig:graphical_abstract}). A unique aspect of our tool is that it allows the user to trace pollutant transport in surface waters from a given watershed or pollutant source through downstream waterbodies and watersheds. Here, we provide a detailed description of the methods used for building a graph representation starting from public and open-source databases such as the  National Hydrography Dataset (NHDPlusV2) \cite{NHDPlus2019} and Watershed Boundary Dataset (WBD) \cite{usgs2019WBD} by using {\tt GeoPandas} \cite{geopandas} and {\tt NetworkX} \cite{NetworkX2008}. 
\\

We demonstrate the capabilities of {\tt HydroGraphs} by providing case studies in the State of Wisconsin, where our resulting graph contains over 45,000 nodes, representing more than 3,000 km\textsuperscript{2} of waterbodies and 79,000 km of rivers and streams. We also provide case studies showing how the graph framework can be applied to analyze upstream sources and downstream impacts of anthropogenic pollution. In these studies, we analyze nutrient pollution in Wisconsin, a challenge that has lasted for decades and that originates from intensive agricultural practices and other anthropogenic sources.  We emphasize with these studies how {\tt HydroGraphs} can easily incorporate data for point and non-point pollutant sources as well as impact data; in particular, we use data for hundreds of concentrated animal feeding operations (CAFOs) as point sources within the graph and include over 93,000 km\textsuperscript{2} of agricultural land as non-point sources of pollution and we show how to link such source data to impact data (chlorophyll-a concentration in lakes to monitor the onset of HABs). With this, we aim to show how {\tt HydroGraphs} can be a valuable tool for researchers and decision-makers to conduct quick assessments of pollutant impacts on the environment. Moreover, we discuss how the framework can be used in conjunction with supply chain optimization models to understand how changes in agricultural practices or in infrastructure can increase (or decrease) the quality of hydrological systems.

\section{Graph Representation of Hydrological Systems}

The focus of our work was to create a graph modeling tool, {\tt HydroGraphs}, to link upstream pollutant sources and downstream pollutant impacts. Specifically, we aim to develop a tool that help us answer questions such as: What waterbodies will be affected by a pollutant release in a specific watershed? What upstream pollutant releases may be impacting a given waterbody? What waterbodies may be $``$storing$"$ pollutants along a given pathway? To answer such questions, it is necessary to model the connectivity between the objects of interest (e.g., pollutant sources or waterbodies) and creating simple and intuitive ways of analyzing these interconnections. Graphs provide a natural mathematical representation to achieve these goals. In this section, we provide an overview on graph representations, outline the data needed to build such graphs, and outline the specific steps to express the graph connectivity. Our framework can be used to capture diverse hydrological systems in the United States (or in the world, provided that data is available in the required format). In the next section, we illustrate these general capabilities by building graph representations to capture interconnectivity of surface waters in the State of Wisconsin and we explore pollutant tracing applications.
\\

\begin{figure}[!htp]
  \centering
  \includegraphics[scale=.65]{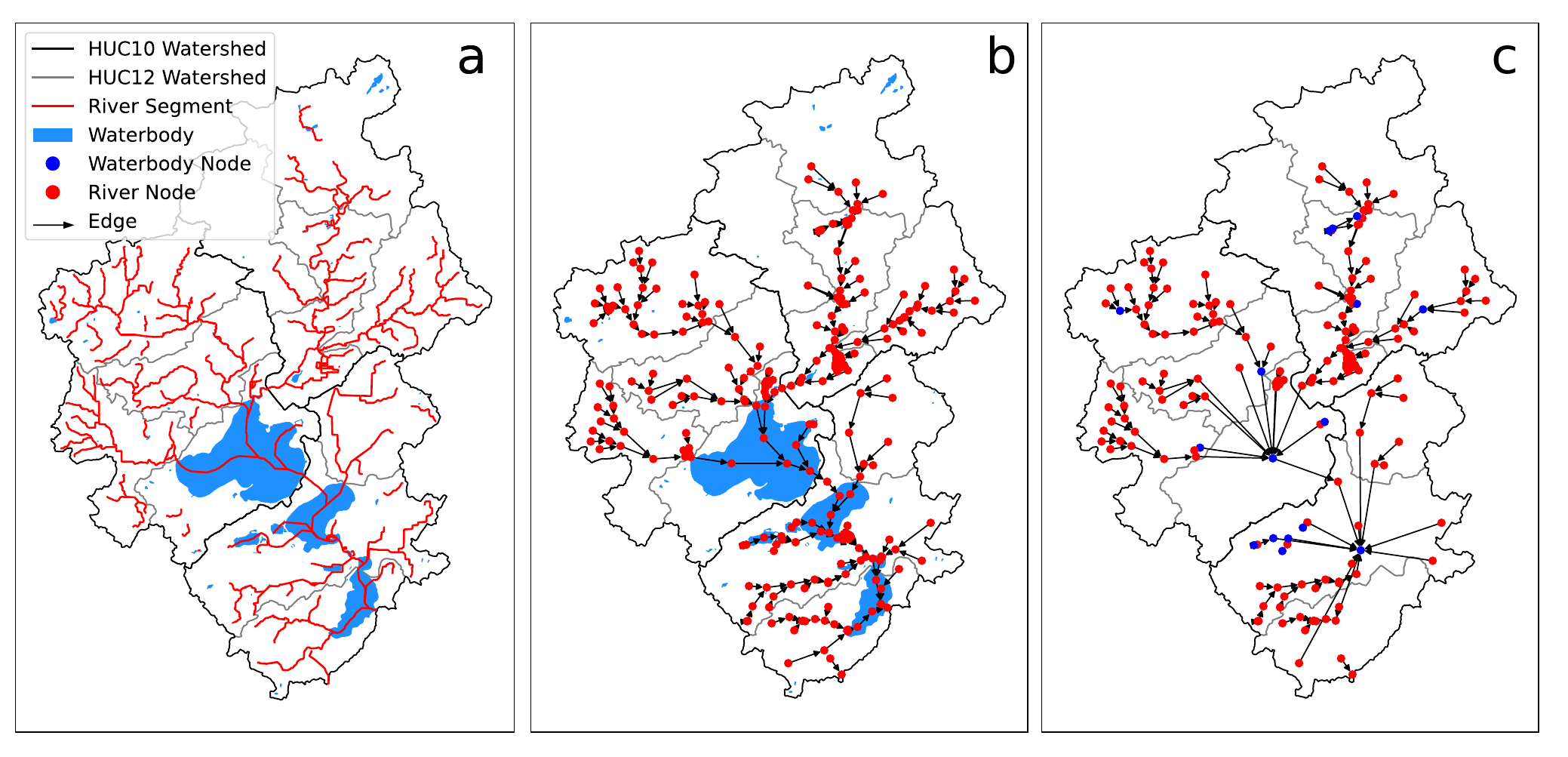}
  \caption{Visualization of methods used for building a graph from geospatial data. We start with river segments and waterbodies from the NHDPlusV2 dataset (panel a) and form a directed graph of the river systems (panel b). We then identify lake connectivity to the river graph and add waterbody nodes to the directed graph (Panel c). Watersheds shown are the HUC10 watersheds 0709000205, 0709000206, and 0709000207, primarily in Dane County, Wisconsin.}
  \label{fig:graph_framework}
\end{figure}

An illustration of the methodology followed by our framework is provided in Figure \ref{fig:graph_framework}. We start with river segments (line features) and waterbodies (polygon features) within watersheds (panel a). From existing data in the NHDPlusV2 (see overview in Section \ref{sec:data_overview}), we create a graph of the river network (panel b) which does not include any waterbodies. We then determine the connectivity of the waterbodies and add them to the river graph (panel c).  

\subsection{Graph Theory Overview}
Graphs are modeling abstractions that are comprised of a set of nodes and edges. Nodes are used to represent diverse objects/elements of a system while edges are used to model connectivity between nodes. As a graph is a mathematical model, there are many different ways in which nodes and edges can be defined and a given selection is often driven by the insights needed from the model and from the data available. To build the watershed-river-waterbody system of interest as a graph, we chose to represent river segments and waterbodies as nodes. Edges are then placed between river segments and waterbodies that flow into one another through surface waters (e.g., by rivers or streams). 
\\

Nodes and edges of graphs can contain attributes (data) that are useful in manipulating and visualizing a graph. For example, an attribute that we used in our model is the watershed in which the node resides (the node encodes spatial/geographical context); this attribute can be used to filter out nodes that lie on a specific watershed. Our representation uses a directed (rather than undirected) graph. In undirected graphs, edges only capture  connectivity (with no notion of directionality); while, in directed graphs, edges capture directionality. In our context, we are interested in tracing nutrient pollution, and we thus need to capture flow directionality. 
\\ 

A key benefit of using graph representations is that there are a wide range of theory and computational techniques for analyzing large-scale graphs \cite{lesne2006complex, quinn1984parallel}; for instance, one can use algorithms to identify a set of nodes that is connected to a given node by using pathway analysis \cite{dijkstra2022note, lee1961algorithm}. Moreover, it is possible to visualize, aggregate, and partition graphs to gain insights into the connectivity and properties of a graph. In addition, it is possible to compute statistics of a given graph object, such as the number of nodes in a given pathway, the fractal dimension of a graph (e.g., a measure of complexity), or the node degree distribution (e.g., number of connections of a node) \cite{even2011graph}. Many tools for graph-based analysis are not readily applicable to other modeling abstractions (e.g., spectral graph theory \cite{spielman2007spectral}, topological data analysis \cite{leykam2023topological, smith2021euler}, community detection \cite{fortunato2010community,palla2005uncovering}, or machine learning approaches like graph convolutional neural networks \cite{zhang2019graph}), potentially providing unique insights into hydrological systems that would not be possible under other tools. Lastly, another key benefit of using graph representations is that there are a wide range of open-source tools that can be leveraged for building and visualizing graphs \cite{jarukasemratana2013recent}. 

\subsection{Data Overview}\label{sec:data_overview}

The graph representation was constructed directly using geospatial data from the NHDPlusV2 and the WBD datasets. These datasets contain geospatial data corresponding to waterbodies, rivers, and watersheds for geographic regions. They also contain data corresponding to the represented objects, such as area of waterbodies/watersheds, downstream length of streams, or waterbody type. While the data within these datasets are for the United States, similar methods could be applied to other geographical areas provided that the data is in similar formats as what is discussed below. We used {\tt GeoPandas} \cite{geopandas} in Python to work with the geospatial data, and we use {\tt NetworkX} \cite{NetworkX2008} for building and managing the graph. The code for our framework is available at \url{https://github.com/zavalab/ML/tree/master/HydroGraphs}. This repository also contains links to the datasets used in this analysis and a script for downloading these datasets.
\\

In our representation there are three main types of objects used to build the graph (see Figure \ref{fig:graph_framework}a): rivers (NHDFlowline from NHDPlusV2; represented by line features), waterbodies (NHDWaterbody from NHDPlusV2; represented by polygon features), and watersheds (from WBD; represented by polygon features). Each of these lines or polygons has a geographic location (e.g., edges of the polygons were represented by specific geographic coordinates) and has a unique identifier. The NHDPlusV2 dataset uses a unique common identifier (COMID) for every individual river segment or waterbody, while the WBD uses a unique hydrologic unit code (HUC) for individual watersheds. In addition, the WBD has different hierarchical levels, with 8-digit, 10-digit, and 12-digit codes depending on the size of the watersheds, where the higher digit codes are partitions of the lower digit codes (e.g., each HUC10 watershed is made up of multiple, smaller HUC12 watersheds). We will use these identifiers in talking about their corresponding objects; for example, we will form nodes out of each river segment or waterbody, and we will identify these nodes by their corresponding COMID. 

\subsection{Expressing Connectivity}

One of the key technical challenges in building the graph representation is identifying the connectivity between the rivers, waterbodies, and watersheds. Conveniently, the NHDPlusV2 dataset provides connectivity between river segments by giving directed pairs of river segments identified by their COMID (i.e., these are given as pairs of $``$from$"$ COMIDs and their corresponding $``$to$"$ COMIDs). This is essentially a list of directed edges of a graph and could be used to build a graph of the river system. In this case, each river segment corresponds to a node, where the node is identified by the river segment's COMID. The graph formed by this list of directed edges (which we will refer to as the "river graph"; see Figure \ref{fig:graph_framework}b) was a basis for building our overall graph. Note, however, that the river graph does not include specific nodes that correspond to waterbodies ($``$waterbody nodes$"$ will be added in a later process). The NHDPlusV2 data overlaps river segments with waterbody polygons, and many polygons may overlap with several river segments. Furthermore, some river segments may overlap with multiple waterbodies, making it difficult to identify which waterbodies flow into one another. Our methods for identifying which river segments overlapped with waterbodies and for adding waterbodies to the given river segment graph are outlined in this subsection.\\

The first step in adding the waterbodies to the graph was identifying the connectivity of waterbodies and rivers. We first built empty lists for every river segment and waterbody polygon. These lists would contain the COMIDs of all intersecting rivers (for waterbodies) or all intersecting waterbodies (for river segments). We then tested every river segment against every waterbody to see if the river segment intersected the waterbody polygon. This was performed within two loops using the GeoPandas' {\tt intersects} function. If a river segment intersected a waterbody, the river segment COMID was added to the waterbody's list of intersecting rivers, and the waterbody COMID was added to the river segment's list of intersecting lakes. These lists could then be used to add the respective lakes to the graph.\\

\begin{figure}[!htp]
  \centering
  \includegraphics[scale=.4]{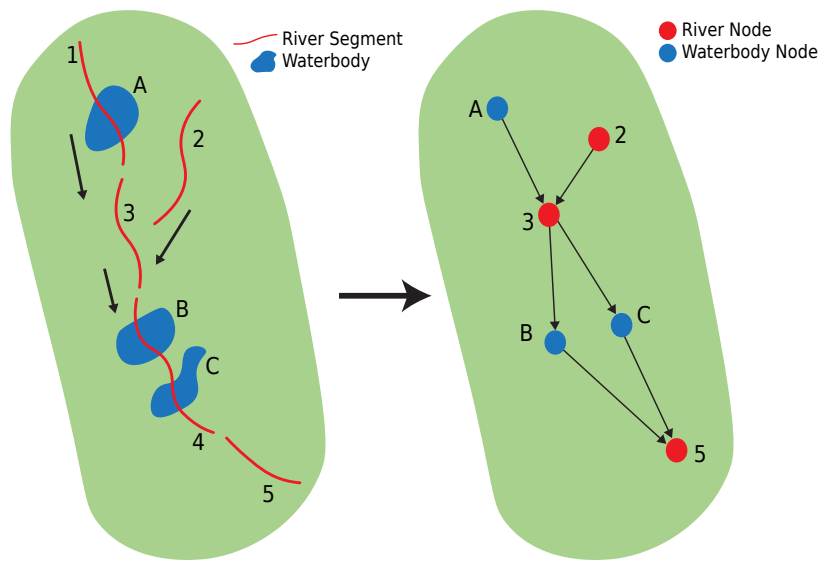}
  \caption{Visualization of methods used for representing the river-waterbody system as a graph. The red lines represent river segments and are identified by numbers while the blue polygons represent waterbodies and are identified with letters. River segment 1 only intersects waterbody A, so it is replaced by a single node corresponding to waterbody A. River segment 4 intersects two waterbodies, B and C, so it is replaced by two nodes within the graph corresponding to waterbodies B and C. Nodes are colored by the object which they represent, where river segments which intersect waterbodies are replaced by a node representing the given waterbody.}
  \label{fig:shapes_to_nodes}
\end{figure}

\begin{landscape}

\begin{figure}[!htp]
  \centering
  \includegraphics[scale=.1]{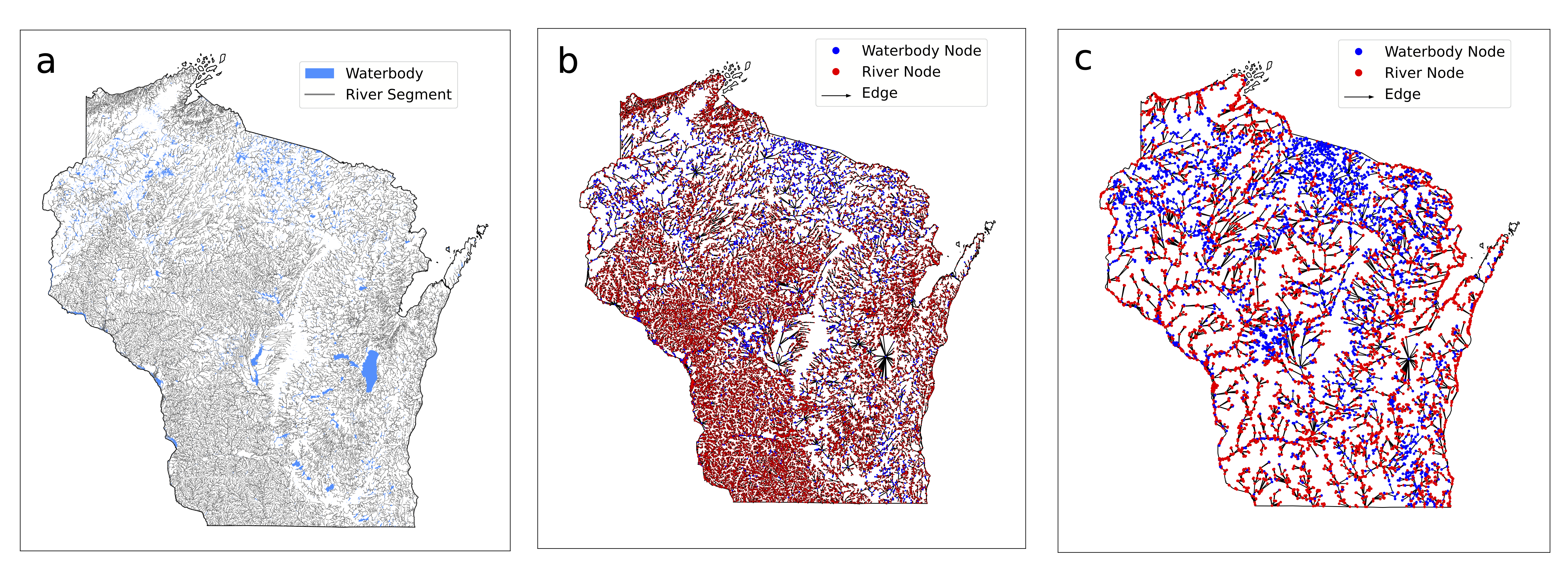}
  \caption{Wisconsin hydrological system (a), representation as a directed graph (b), and aggregated form of the directed graph (c; some river nodes are aggregated to reduce the total number of nodes while maintaining connectivity of waterbodies; see Section \ref{sec:aggregation}). The full directed graph contains over 45,000 nodes and 47,000 edges, representing more than 3,000 km\textsuperscript{2} of waterbodies and 79,000 km of rivers or streams.}
  \label{fig:WI_graph}
\end{figure}

\end{landscape}

After identifying all intersections between rivers and waterbodies, we added waterbodies to the river graph by replacing river COMIDs with waterbody COMIDs (i.e., replacing river nodes with waterbody nodes) and adding edges between the resulting waterbody nodes and other river nodes (See Figure \ref{fig:graph_framework}c). We first looped through every river segment; if the list of intersecting waterbodies for a given river segment contained only one waterbody, we replaced the river segment COMID in the river graph's edge list with the intersecting waterbody COMID. This added several waterbodies to the river graph. After completing this loop, we looped over every river segment again. If the river segment intersected multiple waterbodies, then the river segment was replaced by all waterbodies which it intersected. This meant that more nodes were added to the graph. A visualization of this process is given in Figure \ref{fig:shapes_to_nodes}. Note that this results in waterbodies that intersect the same river segment not necessarily being directly connected (e.g., waterbody B in Figure \ref{fig:shapes_to_nodes} should connect to waterbody C, but it does not). This simplification was made because flow directions in this framework are identified by the FROMCOMID and TOMCOMID combinations and not within the individual flowlines themselves. Thus, it is difficult to identify in an automated way which waterbody is intersected first by a single river segment. We believe that our simplification is reasonable because the general connectivity of the full graph is still maintained, and this simplification is only made on a local scale (flowlines are comparatively short, so this is a local impact, while larger-scale connectivity is maintained). Moreover, this highlights how the graph representation used is inherently limited by the availability of data. The affected waterbodies still have the same upstream and downstream connections with the minor exception of not being connected to waterbody(ies) that intersect their same line segment. For example, waterbody A still flows into waterbodies B and C in Figure \ref{fig:shapes_to_nodes} even though waterbodies B and C are not directly connected to each other. In other words, waterbody A's downstream graph (and river node 5's upstream graph) includes the same set of nodes as it would if this simplification was not applied. For the State of Wisconsin, this latter simplification applied to less than 10\% of the waterbodies in the graph. This simplification is very localized (average river segment length is $<$ 2 km, and the impacts are only on waterbodies connected by the same river segment), and thus we believe this will produce minimal error. Further details about this method can be found in the supporting information. The resulting graph obtained for Wisconsin is presented in Figure \ref{fig:WI_graph}.\\

We make a few additional remarks about our data and methods. First, we chose to exclude a few waterbodies from our graph. We excluded Lake Michigan and Lake Superior from the graph to make the visualizations simpler. Ultimately, for the Wisconsin area explored in the next section, everything that does not flow into the Mississippi River flows into these two lakes (i.e., all nodes in the 04 HUC2 watershed are connected to the Great Lakes), so the connectivity to these lakes is established by nodes being within the 04 HUC2 watershed. Representing either of these lakes by a single node makes the visualizations more difficult to follow, and thus were undesirable for our needs. However, these could be included in our methods simply by not removing the COMIDs for these two lakes from the original dataset. \\

In addition, while swamps and marshes are included in the NHDWaterbody dataset, we removed these objects to simplify the analysis. Swamps and marshes impact nutrient transport differently than some other waterbodies, so we did not want to include them in the same category with lakes and reservoirs. However, we note that swamps and marshes could be influential in nutrient transport (in fact they can be used to control nutrient pollution) \cite{dolph2019phosphorus,fisher2004wetland,verhoeven2006regional,walton2020wetland}. Thus, these will be a subject of future research, but they are outside the scope of this study. Wetlands like swamps and marshes can introduce significant complexity because the pollutant transport can be dependent on soil type and vegetative processes (as is the case for nutrient pollution \cite{fisher2004wetland,walton2020wetland}), and they could significantly impact the time scales of the pollutant transport. This study does not focus on the temporal aspect of hydrological pollutant transport, but rather on the pathways and fate of the pollutants in hydrological systems. \\

We also note that much of the connectivity we give here could be elucidated from the LAGOS-US NETWORKS v1 data set \cite{king2021graph}. The LAGOS-US NETWORKS v1 dataset includes information on the lakes that are connected to the river graph edge list (see {\tt nets\_flow\_medres.csv} \cite{lagosnetworks2021}) by indicating whether an edge of the river graph also goes to or from a lake in the dataset. However, we chose to build our connectivity list from scratch because we wanted to include several waterbodies that were not included in the LAGOS-US NETWORKS v1 dataset. For example, they omit lakes that are $<1$ hectare in size, and they do not include reservoirs (NHDWaterbodies attribute FTYPE equal to $``$Reservoir$"$). Both of these sets of waterbodies could be areas that anthropogenic pollutants accumulate and could be a focus of pollutant studies (see for example \cite{wang2018heavy,oliver2019challenges}). Excluding these waterbodies could thus lead to incorrect results when seeking to identify areas of pollutant impacts.\\ 

We recognize that the above methods are only focused on waterbodies that are connected through surface waters. The resulting graph outlined above does not include every waterbody in a geographic area because many waterbodies are isolated and not connected by surface waters to other objects in the graph. Furthermore, this graph is specific to surface water and does not include transport through other means such as groundwater, diffusion, or other forms of transport which could be important factors \cite{meinikmann2015phosphorus, valiela1990transport,wang2015inefficient}. Including other transport mechanisms greatly impacts the complexity and will be explored in future work.

\subsection{Aggregating River Nodes}\label{sec:aggregation}

The above methods for building this graph result in several intermediate river nodes upstream or downstream of waterbodies. In many cases, it may be desirable to aggregate nodes to simplify the model representation, either to make the visualizations simpler or to reduce the number of nodes involved when analyzing the graph with varying degrees of spatial resolution. Our framework provides capabilities for automating aggregation; details of this aggregation procedure are included in the supporting information. 
\\


\section{Wisconsin Case Studies}

We highlight how {\tt HydroGraphs} can be used to identify pollutant sources and their potential destinations; we do this by developing some specific case studies (Table \ref{tab:case_studies}). Case studies focus on nutrient pollution in Wisconsin, a challenge that has existed for decades due in part to the large amount of agricultural land and CAFOs throughout the state that result in nitrogen (N) and phosphorus (P) flowing into nearby waterways. Nutrient losses from these sources frequently lead to HABs, which can have negative health, economic, and environmental impacts for local communities.
\\

\begin{table}[!htp]
    \caption{Overview of the three case studies presented in this work}
    \centering 
    \begin{tabular}{l|l} 
    \hline\hline
    Case Study 1 & Upstream Watershed Analysis\\
    \hline
    Case Study 2 & Metrics for Data Analysis\\
    \hline
    Case Study 3 & Downstream Pathway Analysis\\
    \hline\hline
    \end{tabular}
    \label{tab:case_studies}
\end{table}

The first case study compares a couple of lakes in Wisconsin with differing total phosphorus (TP) concentrations and looks at their upstream graphs and likely P sources that contribute to these differences. The second case study looks more generally at several hundred lakes for which we have TP and chlorophyll-a data and compares connectivity attributes of the graph between polluted and clean lakes. The final case study looks at how our framework can be used to identify impacts that a potential pollutant source could have. This is done by inspecting the nodes in the graph that are downstream of the source.
\\

The case studies presented herein are intended as examples of ways that this graph framework could be applied. While these case studies do incorporate real data, they are not rigorous studies intended to give exact causation or to make policy recommendations. Rather, their purpose is to present how the graph could be applied by experts and researchers in this field. Further, they highlight how {\tt HydroGraphs} could be used in helping decision-makers approach complex problems involving pollutants in hydrological systems. For the code to replicate these case studies, see \url{https://github.com/zavalab/ML/tree/master/HydroGraphs}.

\subsection{Case Study I: Identifying Upstream Sources}

In this case study, we look at how this graph can enable identifying upstream influences to a given waterbody. We focus in this case on P pollution in waterbodies, but the principles in this case study could easily be applied to other pollutants, such as ECs, microplastics, or heavy metals. Here, we build the upstream graphs for two lakes in Wisconsin---Lake Altoona and Mohawksin Lake---and identify possible pollutant sources that contribute to these lakes. We choose these lakes because we have TP concentration measurements from the Wisconsin DNR for each lake \cite{dnraltoona,dnrmohawksin}. Based on the measured TP concentrations and reported lake perception, Lake Altoona has poorer water quality (average measured TP of 103 mg/m\textsuperscript{3}) and has noticeably worse problems with algae than Mohawksin Lake (average measured TP of 40 mg/m\textsuperscript{3}). Data for these lakes and their reported perception measurements are available in the supporting information.\\

There are several upstream factors that can impact pollutant transport to waterbodies. We look at four specific factors that may influence the TP concentrations within the waterbodies. The first factor is the waterbodies that are upstream to a given waterbody. Upstream waterbodies could accumulate pollutants and could impact how much of a pollutant reaches a downstream waterbody and when it reaches it \cite{carpenter2014connectivity, motew2017influence,jones2010connectivity}. These waterbodies are naturally a part of the upstream graph from a specific waterbody. The second factor we consider are CAFOs; these can be a source of pollutants in Wisconsin because of their large production of manure, and many sources show that these can be significant contributors of P to waterbodies \cite{burkholder2007impacts, long2018use,parry1998cafos}. For our analysis here, we use data from Hu and co-workers\cite{hu2018supply} to identify locations for more than 200 CAFOs. We add the CAFOs from this dataset to the directed graph outlined above by adding a directed edge from the location of the CAFO to the closest node in the same HUC12 watershed as the given CAFO. The third source we consider is agricultural land; agricultural land is a significant non-point source of N/P and can be an indicator of surface water pollutant concentrations \cite{carpenter1998nonpoint, le2010eutrophication,motew2019comparing,Robertson2006usgs}. This source can be closely related to CAFOs because the manure from CAFOs is often applied as a nutrient source for crops. We use the shapefile from \cite{james2020agriculture} to identify agricultural land, and we only look at agricultural land that shares a HUC12 watershed with the given waterbody or its upstream nodes. For this dataset of agricultural land, we only use the polygons that are labeled as agricultural land that exclude pasture class (where feature {\tt isAG} equals 1). Lastly, we also consider the urban land cover in the given watershed. Urban land cover can have significant impacts on water quality (see for example \cite{hobbie2017contrasting, yang2020spatial,tasdighi2017relationship}), and we use the data from the Wiscland 2 Land Cover Dataset \cite{dnrwiscland2} produced by the Wisconsin DNR which classifies all land throughout the state, including a subclassification of "urban land" (see the Supporting Information for how this data is curated and used in analysis).\\ 

\begin{figure}[!htp]
  \centering
  \includegraphics[scale=.75]{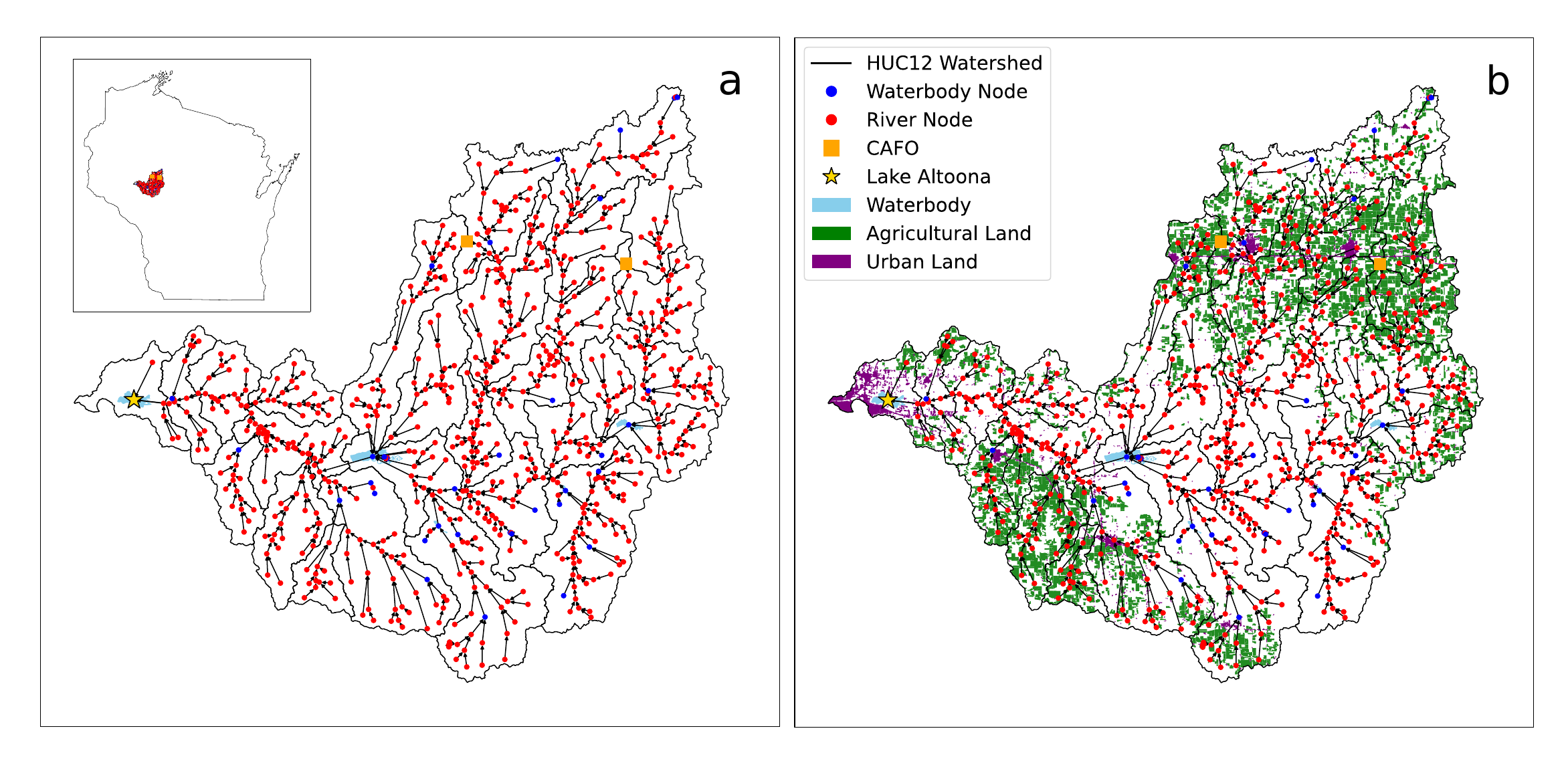}
  \caption{Upstream graph of Altoona Lake in Wisconsin. Panel a) shows all upstream river and waterbody nodes along with CAFOS connected to the closest node within their HUC12 watershed. Panel b) includes agricultural land polygons in green and urban land polygons in purple.}
  \label{fig:altoona_lake}
\end{figure}

\begin{figure}[!htp]
  \centering
  \includegraphics[scale=.7]{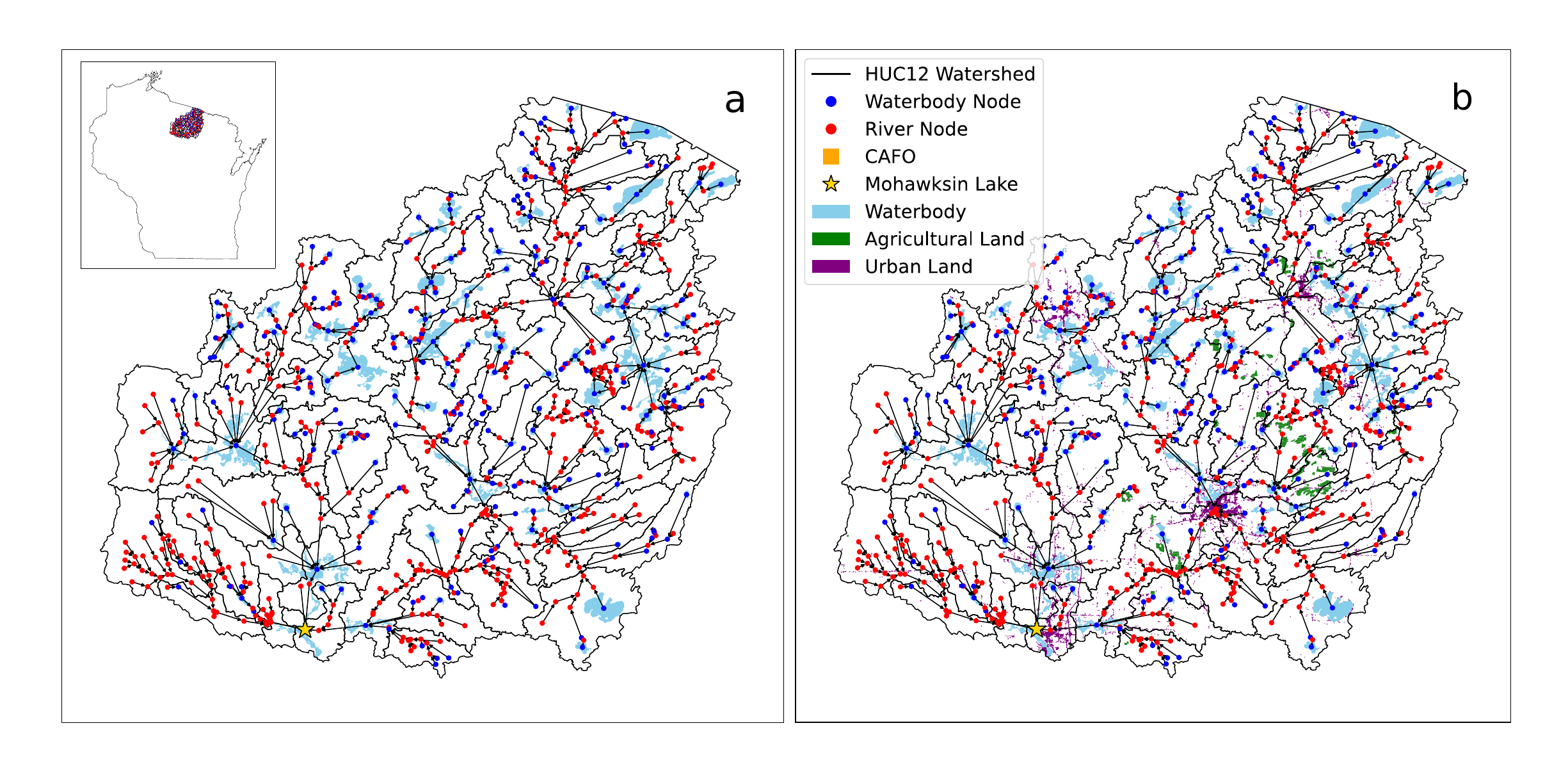}
  \caption{Upstream graph of Lake Mohawksin in Wisconsin. Panel a) shows all upstream river and waterbody nodes. Panel b) includes agricultural land polygons in green and urban land polygons in purple.}
  \label{fig:lake_mohawksin}
\end{figure}

Because we are using a directed graph representation, we can easily form the upstream graphs for both Altoona Lake and Lake Mohawksin. As each node of the graph can be mapped to a single HUC12 watershed, we can identify the watershed corresponding to the upstream graph and thus easily compute agricultural and urban land fractions or add connections of point sources (e.g., CAFOs) to the graph for analysis. The identification of the upstream watershed is made possible by being able to readily compute the upstream graph (and its corresponding nodes) through efficient graph algorithms. This allows us to frame potential hypotheses of why Altoona Lake has much higher TP concentrations than Lake Mohawksin. The upstream graphs for these lakes can be seen in Figures \ref{fig:altoona_lake} and \ref{fig:lake_mohawksin}. Panel a) shows all upstream nodes (including CAFOs) overlayed on the HUC12 watersheds and where these watersheds lie in Wisconsin. Panel b) includes the agricultural and urban land polygons. These figures also include the upstream waterbody polygons to give an idea of the size of the waterbodies to which waterbody nodes correspond. 
\\

From Figures \ref{fig:altoona_lake} and \ref{fig:lake_mohawksin}, it is clear that both Altoona Lake and Lake Mohawksin have large upstream graphs that include several waterbodies. Altoona Lake includes 34 upstream waterbodies that cover an area of more than 7 km\textsuperscript{2}, while Lake Mohawksin includes 265 upstream waterbodies covering more than 345 km\textsuperscript{2}. Altoona Lake also has upstream connections to two CAFOs while Mohawksin Lake has no CAFOs in its upstream watersheds. The figures also show that there is significantly more agricultural land in Altoona Lake's upstream graph. The agricultural land makes up 20.6\% of Altoona Lake upstream watersheds while it makes up only 0.7\% of Lake Mohawksin's upstream watersheds. Further, Altoona Lake had a higher percentage of urban land (2.3\%) than Mohawksin Lake (1.7\%) while also having a denser concentration of urban land around the lake itself. The addition of CAFOs, the higher fraction of agricultural land, and the higher fraction of urban land can thus be likely contributors to the high TP concentrations of Altoona Lake.\\

The above analysis gives some examples of how {\tt HydroGraphs} could be used for analysis. It is very possible that the high amount of agricultural land (agricultural land fraction is more than ten times higher for Lake Altoona as Mohawksin Lake) and the upstream CAFOs (two CAFOs for Lake Altoona compared to zero CAFOs for Mohawksin Lake) are contributors to the high TP concentrations within Altoona Lake. Further, the high number of upstream waterbodies for Lake Mohawksin may influence the transfer of pollutants to Lake Mohawksin (e.g., through the accumulation of nutrient pollution in upstream waterbody sediments) \cite{carpenter2014connectivity,leavitt2006landscape,soranno1999spatial}. Formulating these systems as a graph enables the above visualizations and simplifies analysis. It makes it easier to identify upstream point sources that could contribute to pollutant concentrations in waterbodies. In addition, just as CAFOs were added to the graph, other pollutant sources (such as wastewater treatment plants \cite{brooker2018discrete,  makarewicz2012tributary} or landfills \cite{hu2018supply}) could also be added. This could be useful in the event that a pollutant, such as an EC, is discovered in a given lake or stream. Building the upstream graph would allow researchers and decision makers to identify where this pollutant may be coming from, and to identify other upstream waterbodies or rivers that may need to be tested to see if they are likewise contaminated. As seen from the CAFOs in Figure \ref{fig:altoona_lake}, some of these upstream pollutant sources could be far upstream but are more easily identifiable by building the graph. 

\subsection{Case Study II: Graph Connectivity Metrics}

In this case study, we study upstream graph metrics for waterbodies for which we have TP and chlorophyll-a (a measure that relates to the level of algae in the lake) data from the DNR \cite{dnr}. The DNR provides water quality data collected by volunteers for hundreds of waterbodies in Wisconsin. We compiled their data for more than 700 unique waterbodies. Data for each waterbody could vary in terms of frequency of measurements and type of measurements taken. To ensure that the lakes considered had significant data, we only studied waterbodies for which there were at least 50 measurements for both TP and chlorophyll-a which resulted in a set of 241 waterbodies. We chose the cutoff of 50 to ensure that we had several data of both TP and chlorophyll-a while maintaining a reasonable subset of lakes to analyze (e.g., choosing a higher cutoff such as 100 resulted in too few lakes, while choosing a smaller cutoff like 10 could result in too little data for the lakes). For more details on how this data was compiled and what was included, see the Supporting Information. \\

\begin{landscape}
  \begin{figure}[!htp]
    \centering
    \includegraphics[scale=.8]{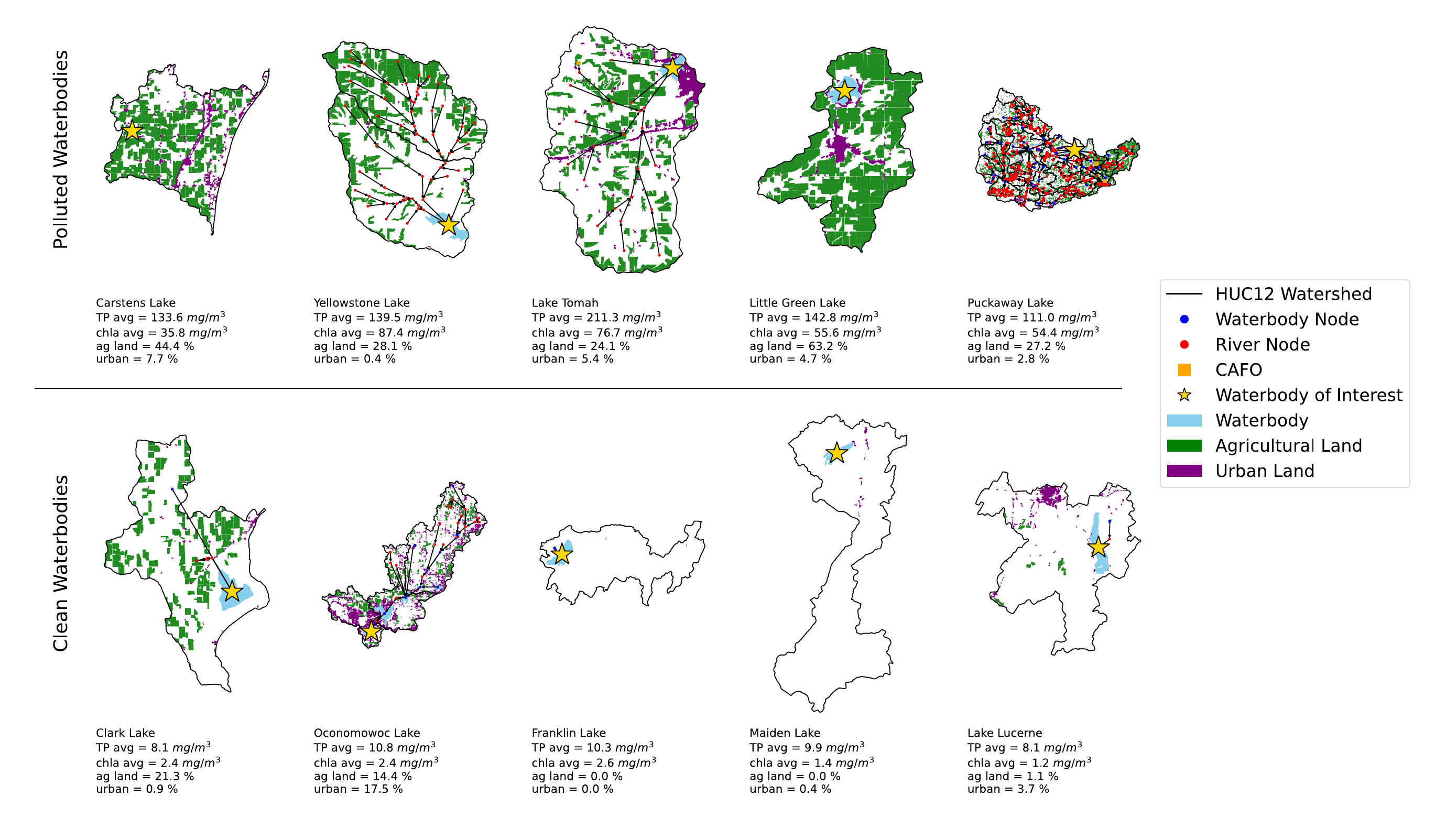}
    \caption{Comparison of upstream graphs for five polluted waterbodies and five clean waterbodies, as determined by data from the Wisconsin DNR \cite{dnr}. "TP avg" and "chla avg" are the average TP and chlorophyll-a measurement, and "ag land" and "urban" are the agricultural and urban land fractions, respectively, in the HUC12 watersheds comprising the graph.}
    \label{fig:10_lakes}
  \end{figure}

\end{landscape}

Based on this data for TP and chlorophyll-a, we studied five waterbodies that had the highest and lowest levels of TP and chlorophyll-a within our graph. The times at which lakes were sampled varied between waterbodies; as such, we averaged all reported measurements of TP and chlorophyll-a. This is a primary reason why we required that lakes have at least 50 data points for both TP and chlorophyll-a as we assumed that this would reduce or eliminate any differences between waterbodies caused by different sampling times. We then looked at the five waterbodies that were in our graph that had the highest average TP measurement and an average chlorophyll-a measurement of at least 30 mg/m\textsuperscript{3}, and we looked at the five waterbodies with the lowest average TP measurement and an average chlorophyll-a measurement of no more than 5 mg/m\textsuperscript{3}. The cutoff values for chlorophyll-a were chosen to ensure that the given waterbodies were at the high or low extremes of both TP and chlorophyll-a within our dataset (for further analysis of and explanation for these cutoff values, please see the Supporting Information). After determining these five most and least polluted waterbodies for which we have data, we built the upstream graphs using similar methods as discussed in Case Study I, where we added CAFOs, agricultural land, and urban land to the visualizations. The results can be seen in Figure \ref{fig:10_lakes}.\\

By creating the graphs in Figure \ref{fig:10_lakes}, we are able to observe some trends between the polluted and clean waterbodies. The polluted waterbodies all had agricultural land fractions higher than the clean waterbodies, and no polluted waterbody had an agricultural land fraction lower than 24\%. Both Lake Tomah and Puckaway Lake (polluted waterbodies) had upstream CAFOs while no clean waterbody had an upstream CAFO. Urban land appears to be a more complicated attribute between waterbodies as the lake with the highest urban land fraction (Oconomowoc Lake) was categorized as a "clean waterbody", though the other clean waterbodies on average had a much lower urban land fraction than the polluted waterbodies. The polluted waterbodies also had three waterbodies with significant upstream connections. In addition, there were two waterbodies in the polluted waterbodies (Carstens Lake and Little Green Lake) and one in the clean waterbodies (Maiden Lake) that had no upstream connections. This is because they are in the graph, but only have downstream connections. \\

The analysis of Figure \ref{fig:10_lakes} can also be expanded to include more waterbodies. To look at how well these metrics apply to other waterbodies, we took a bigger subset of waterbodies by taking as "clean" all waterbodies with TP concentrations of $<$ 15 mg/m\textsuperscript{3} and chlorophyll-a concentrations of $<$ 5 mg/m\textsuperscript{3} which totaled 60 waterbodies. For the "polluted", we used all waterbodies with TP concentrations of $>$ 60 mg/m\textsuperscript{3} and chlorophyll-a concentrations of $>$ 15 mg/m\textsuperscript{3} which totaled 18 waterbodies (for further analysis of and explanation for these cutoff values, please see the Supporting Information). We then looked at how many of these waterbodies are connected to CAFOs upstream (or, in the case of waterbodies without an upstream graph, if there is a CAFO in their HUC12 watershed), what are their agricultural and urban land fractions, how many waterbodies are not in the graph (i.e., how many are not connected to a river or stream in the graph), and how many upstream nodes these waterbodies have that were in the graph. The results of this analysis are shown in Table \ref{tab:graph_metrics}. In addition, we also found that the average agricultural land fraction for the polluted waterbodies (26.7\%) was three times higher than the average for the clean waterbodies (8.0\%), and the average urban land fraction was almost one and a half times higher for the polluted waterbodies (5.55\%) than the average for the clean waterbodies (3.80\%).\\

\begin{table}[!htp]
    \caption{Metrics for Polluted and Clean waterbodies using {\tt HydroGraphs}. Metrics indicate the number of waterbodies that are connected upstream to a CAFO (CAFO); the number of waterbodies that are in the graph (In Graph); the number of waterbodies that could be considered headwaters as they are in the graph but have no upstream nodes (Headwater); the number of waterbodies that have an agricultural land fraction of more than 0.2 in their watershed ($>$ 20\% ag); the number of waterbodies with an urban land fraction of more than 0.02 ($>$ 2\% urban); and the number of waterbodies that have at least 10 nodes in their upstream graph (10$+$ nodes).}
    \centering 
    \begin{tabular}{c cccccccc} 
    \hline\hline 
     &  & Total & CAFO & In Graph & Headwater & $>$ 20\% ag & $>$ 2\% urban & 10$+$ nodes \\
    \hline
    \hline 
    \multirow{2}{*}{\shortstack{Polluted \\Waterbodies}}& Number & 18 & 6 & 15& 2& 13 & 12 & 9\\ 
     & Fraction &  &0.33  &0.83 &0.11 &0.72 & 0.80 & 0.50\\
  
    \hline
    \multirow{2}{*}{\shortstack{Clean \\Waterbodies}}& Number & 60 & 4 & 24& 10 & 9 & 11 & 3\\ 
    & Fraction &  &0.07& 0.40& 0.17& 0.17 & 0.46 &0.05 \\
   \hline
    \end{tabular}
    \label{tab:graph_metrics}
\end{table}

Overall, this analysis suggests that some graph metrics may be feasible indicators that help identify polluted lakes. One third of the polluted waterbodies were connected to an upstream CAFO, and the polluted waterbodies were generally in watersheds with much higher agricultural land fractions and (on average) higher urban land fractions. Further, the polluted waterbodies often exhibited much higher upstream connectivity than the clean waterbodies. These metrics are all relatively easy to compute using the graph representation of the hydrological system, and the results could potentially be extrapolated to other lakes for which we do not have data compiled. Building these systems as a graph provides new tools for pollutant fate and transport. We would like to highlight that the results found in this study are only speculative and do not aim to provide a final recommendation on the origins of Lake nutrient pollution (which can be the result of many factors). 

\subsection{Case Study III: Downstream Impacts}

\begin{figure}[!htp]
  \centering
  \includegraphics[scale=.55]{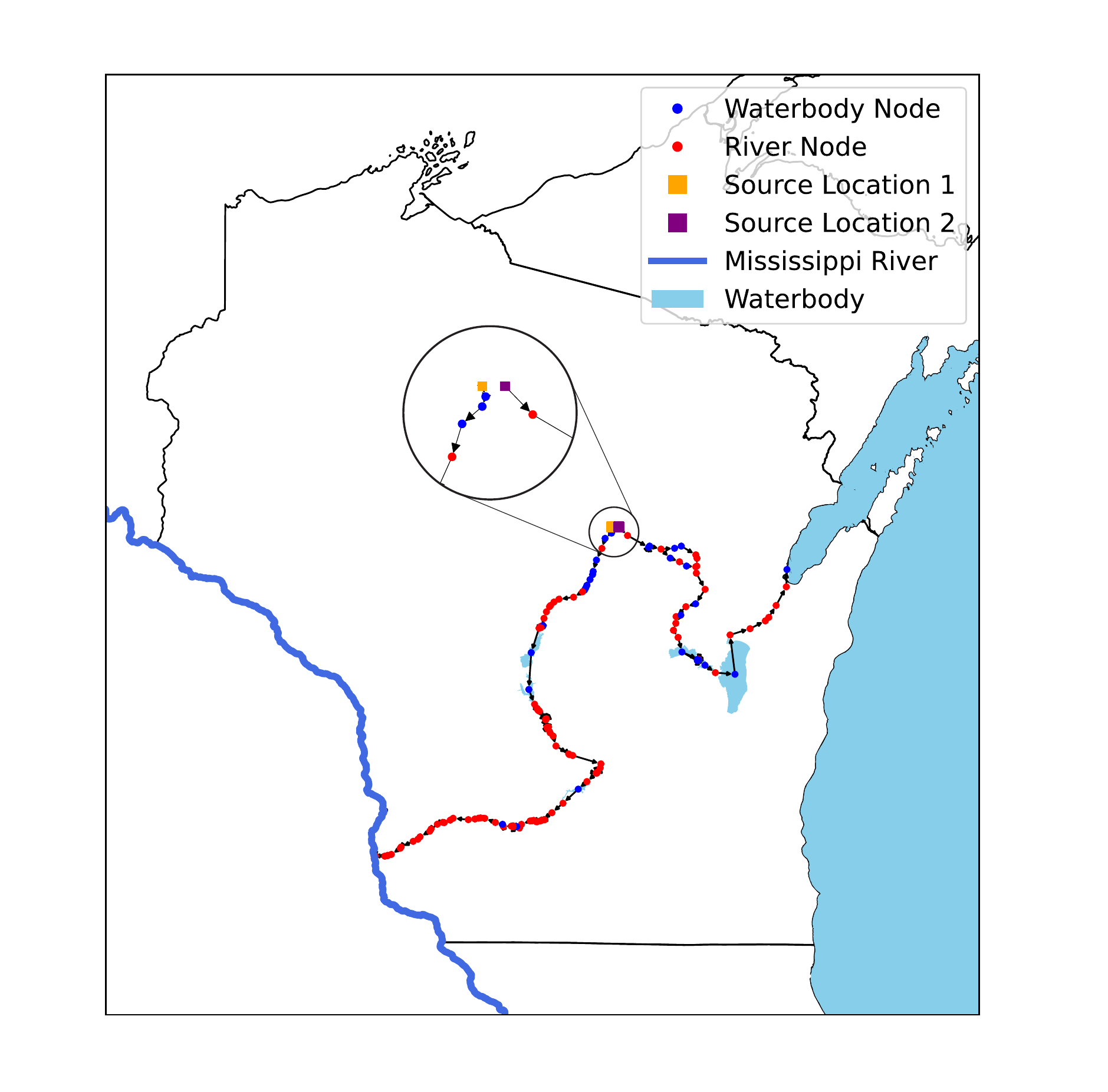}
  \caption{Downstream aggregated graphs for two potential pollutant sources. Source locations are separated by less than 5 km}
  \label{fig:downstream_graphs}
\end{figure}

For this final case study, we consider a hypothetical example of placing a potential pollutant source (e.g., a new CAFO or wastewater treatment plant) and identifying its downstream impacts. In determining where to place this new pollutant source, we need to consider the potential downstream impacts this could have. We look at a couple of potential locations for its placement within the same area. Interestingly, the locations are just within five kilometers of each other, but they have drastically different downstream destinations (because they lie in different watersheds). These pollutant sources are added to the graph by placing a directed edge from the pollutant source to the nearest node within the pollutant source's HUC12 watershed. The downstream, aggregated graphs for these two pollutant sources are shown in Figure \ref{fig:downstream_graphs}.\\

Location 1 (Figure \ref{fig:downstream_graphs}a) results in the pollutant passing through 15 waterbodies and going into the Mississippi River (the Western border of Wisconsin), while location 2 (Figure \ref{fig:downstream_graphs}b) shows the pollutant also passing through 15 waterbodies and moving into Lake Michigan. This analysis can be useful in identifying waterbodies potentially impacted by new pollutant releases. In the event of a contaminant release, this graph can help identify waterbodies that may be impacted by such release and this can help develop mitigation/response strategies. Furthermore, the graph can also be useful for decision-makers in identifying where to place potential pollutant sources (such as building a new wastewater treatment plant) to minimize that sources impacts on the environment and communities. The locations shown in Figure \ref{fig:downstream_graphs} travel through completely different waterbodies, and these waterbodies may be at different stages of eutrophication and may be of varying importance/priority (e.g., if some waterbodies serve as drinking water for a locality, it may be more important than another waterbody). Thus, building the downstream graph enables researchers or decision-makers to see potential impacts of a newly introduced pollutant source.

\subsection{Conclusions and Future Work}

{\tt HydroGraphs} provides a framework for analyzing hydrological pollution pathways to identify upstream sources and downstream impacts. The above case studies have shown how {\tt HydroGraphs} can be used with point and non-point pollution sources to identify upstream sources and link attributes within the graph to pollutant data. Further, it can also help identify potential downstream impacts from a given pollutant source. While the case studies in this paper focused on anthropogenic nutrient pollution, similar methods could be applied for other pollutants such as ECs or microplastics. Point or non-point sources of these contaminants could be added to the graph following a similar analysis as that done with CAFOs, agricultural land, and urban land. Building these hydrological systems as a graph ultimately provides simple visualization and rapid analysis.\\

There are two areas we would like to address in the future using {\tt HydroGraphs}. First, there is additional data that we can incorporate into the graph. For example, the NHDPlusV2 dataset includes attributes such as average stream flowrates that could be added to the graph as edge weights. As waterbodies often have multiple streams flowing into them, these flowrates could show which upstream sources have a stronger impact on the waterbody (e.g., the upstream sources connected through the larger stream may have a stronger impact). Second, we would like to incorporate {\tt HydroGraphs} into additional decision-making models, such as into supply chain optimization. For example, Tominac et al. \cite{tominac2020evaluating} included environmental policy-makers as stakeholders within their supply chain model. One of the challenges in doing so is quantifying the environmental or social impacts within the supply chain. {\tt HydroGraphs} could provide a tool for quantifying these impacts, such as quantifying the number of lakes that would be impacted by the introduction of a new pollutant source. This would allow for these decision-making models to highlight the environmental, economic, and social impacts of many pollutant sources.


\section*{Supporting Information}
Additional methodological details on graph construction and aggregation, details of the DNR data presented herein, and an overview of functionality for working with the graph representation are provided in the SI. 


\section*{Acknowledgments}

We acknowledge support from the U.S. EPA (contract number EP-18-C-000016).  We thank Eric Booth for helpful feedback on an early version of this manuscript. 
\\

The views expressed in this article are those of the authors and do not necessarily reflect the
views or policies of the U.S. Environmental Protection Agency. Mention of trade names, products,
or services does not convey, and should not be interpreted as conveying, official U.S. EPA approval, endorsement, or recommendation.

\bibliography{WisconsinGraph}

\end{document}



\title{{\bf Supplementary Information}\\ A Graph-Based Modeling Framework for Tracing \\ Hydrological Pollutant Transport in Surface Waters}

\author{David L. Cole${}^{a}$, Gerardo J. Ruiz-Mercado${}^{bc}$, and Victor M. Zavala${}^{a}$\thanks{Corresponding Author: victor.zavala@wisc.edu}
}

\date{\small
  ${}^a$Department of Chemical and Biological Engineering, \\[0in]
  University of Wisconsin-Madison, Madison, WI 53706 \\
  ${}^b$Office of Research and Development, \\
  U.S. Environmental Protection Agency, Cincinnati, OH 45268, USA,\\
  and Chemical Engineering Graduate Program, \\
  Universidad del Atl\'antico, Puerto Colombia 080007, Colombia
}

\maketitle
\pagebreak
\tableofcontents

\section{Graph Construction Methods}
In this section, we give further details and methodology for our graph construction. Code for building our graph for Wisconsin can be found at \url{https://github.com/zavalab/ML/tree/master/HydroGraphs}. The code for building this graph is found in the "graph\_construction" folder in the above link. We also note that our methods ultimately result in a list of edges for the graph, where each entry of the list is a set of two nodes. We ultimately store this list in the format of the National Hydrography Dataset's (NHDPlusV2) \cite{NHDPlus2019} PlusFlow data. This data contains a "FROMCOMID" column and a "TOCOMID" column, where the "COMID" is a common identifier for river and waterbody objects. The FROMCOMID is the originating node of the directed edge and the TOCOMID is the receiving node of the directed edge. This list can be turned into a graph using functions we define that access tools within the Python package {\tt NetworkX} \cite{NetworkX2008}, as will be shown in section \ref{sec:functions}. In addition, we also form two GeoDataFrames from {\tt GeoPandas} \cite{geopandas}. One of these GeoDataFrames contains all of the waterbodies in the area of interest and the other contains all of the rivers. Attributes (such as the HUC12 code) of a given waterbody or river is stored within these GeoDataFrames. Because these GeoDataFrames have a column for the COMID, node attributes of the graph can be stored within these GeoDataFrames by adding a column for that attribute and setting a value in the row that corresponds to a given COMID.\\

Before running the code to build the graph, we compiled the required data and shapefiles. Our framework requires the HUC8, HUC10, and HUC12 watershed shapefiles obtained from the Watershed Boundary Dataset (WBD), which we downloaded from \cite{usgs2019WBD} . We also downloaded the river and waterbody shapefiles (NHDFlowline and NHDWaterbody) from the NHDPlusV2 downloaded from \cite{nhdplusdownload}. Because we wanted to build the framework for the state of Wisconsin, we included the shapefiles for both the 04 and 07 HUC2 watersheds. In addition, we included the PlusFlow data from the NHDPlusV2 dataset which contains the "TOCOMID" and "FROMCOMID" data for the NHDFlowline. We converted the file PlusFlow.dbf into a .csv for use in our framework. We also built a general shapefile for the state of Wisconsin. We did this by starting with a shapefile of all Wisconsin counties \cite{dnr_counties} and dissolving these into a single polygon. This resulting polygon was then used to get the lakes, rivers, and watersheds that applied to our area. All of these starting files can be found in the folder "graph\_construction/lakes\_rivers".\\

After assembling this initial data, we built shapefiles containing the geographic area of interest for all of the watersheds, rivers, and waterbodies. This was done by overlaying the polygon of Wisconsin with all other shapefiles. The resulting shapefiles were saved to the folder "WIgeodataframes." The methods for doing this can be found within the file "build\_base\_dataframes.py" where we convert all shapefiles to the same crs code and overlay using {\tt GeoPandas}. In addition, we also add HUC8 codes to all HUC10 shapefiles and we add HUC8 and HUC10 codes to all HUC12 shapefiles.\\

After building the shapefiles for our area of interest, we also added additional data to our waterbody and river GeoDataFrames. We first add the HUC8, HUC10, and HUC12 codes to all rivers and waterbodies. This was done by getting the centroid of the river or waterbody object and finding within what HUC polygon that centroid lies. After completing this (see "add\_hucs\_to\_lakes\_rivers.py"), we then looped through these objects and added the list of intersecting rivers (for the waterbody objects) or the list of intersecting waterbodies (for the river objects). These lists were later used in adding the waterbodies to the river graph. This process was performed within the script "add\_river\_lake\_nodes.py".\\

After adding the river and waterbody intersection columns to our dataframes, we then built the list of edges for our graph. We do this by starting with the PlusFlow.csv list containing the FROMCOMID and TOCOMID list for the rivers. We then add waterbodies to this list using the methods described in our manuscript. This was done within "add\_to\_comid\_list.py". Finally, we include an optional aggregation step "aggregate\_graph.py" which we outline in the next section. This above process results in a list of edges (FROMCOMID and TOCOMID lists) that can be converted to a graph.

\section{Graph Aggregation}

In this section, we present the algorithm we used to aggregate our graph. "Aggregation" involves combining nodes (or sets of nodes) into a single node. Aggregating the graph can be useful to simplify visualizations and to decrease the number of nodes in the graph. This latter point can be helpful in decreasing computation time for different graph algorithms because there are less nodes to iterate through. Our algorithm aggregates some river nodes together and it aggregates some river nodes into waterbody nodes. Aggregation only occurs if certain conditions are met and if two nodes are in the same HUC12 watershed. Within our algorithm, aggregation never merges two waterbody nodes together, thus making sure that the total number of waterbody nodes in the graph remains the same before and after aggregation. This also makes sure that the connectivity of those waterbody nodes does not change.\\

Our algorithm for performing this aggregation is an iterative process. It iterates through the set of all edges in the graph (by iterating through the list of TOCOMIDs and FROMCOMIDs), tests certain conditions, and merges nodes if those conditions are met. After all edges have been passed through, the process repeates until no nodes can be merged. The code for performing this aggregation can be found in "aggregate\_graph.py" and "WI\_graph\_functions.py" at the Github link in the above section. \\

The algorithm iterates throught he list of TOCOMIDs and FROMCOMIDs (i.e., the edges of the graph), and goes as follows:

\begin{itemize}
  \item If both the nodes of the edge are waterbodies, then do nothing to the edge.
  \item If the FROMCOMID node is a waterbody and the TOCOMID node is a river, test if the TOCOMID river node has any immediate upstream connections besides the FROMCOMID waterbody node. 
  \begin{itemize}
    \item If there are no other immediate upstream connections, test if the FROMCOMID waterbody node and the TOCOMID river node are in the same HUC12 watershed. 
    \begin{itemize}
      \item If they are in the same HUC12 watershed, merge the TOCOMID river node with the FROMCOMID waterbody node. Replace all instances of the merged river COMID in the list of TOCOMIDs and FROMCOMIDs with the waterbody COMID.
      \item If they are not in the same HUC12 watershed, do nothing to the edge and move on to the next edge in the list.
    \end{itemize}
    \item If there are other immediate upstream connections besides the FROMCOMID waterbody node, do nothing to the edge and move on to the next edge in the list.
  \end{itemize}
  \item If the FROMCOMID node is a river and the TOCOMID node is a waterbody, test if the FROMCOMID river node has any immediate downstream connections besides the TOCOMID waterbody node.
  \begin{itemize}
    \item If there are no other immediate downstream connections, test if the FROMCOMID river node and the TOCOMID waterbody node are in the same HUC12 watershed. 
    \begin{itemize}
      \item If they are in the same HUC12 watershed, merge the FROMCOMID river node with the TOCOMID waterbody node. Replace all instances of the merged river COMID in the list of TOCOMIDs and FROMCOMIDs with the waterbody COMID.
      \item If they are not in the same HUC12 watershed, do nothing to the edge and move on to the next edge in the list.
    \end{itemize}
    \item If there are other immediate downstream connections, do nothing to the edge and move on to the next edge in the list. 
    
  \end{itemize}
  \item If both the FROMCOMID and TOCOMID nodes are river nodes, then test if there is another immediate upstream node flowing into the TOCOMID river node. 
  \begin{itemize}
    \item If there is another immediate upstream node(s), test if the upstream nodes are connected immediately downstream to any nodes other than the TOCOMID river node. If they are connected to other nodes, do nothing to the edge and move on to the next edge in the list.
    \item  If there is another immediate upstream node(s), test if any of them are waterbodies. 
    \begin{itemize}
      \item If any of the immediate upstream nodes are waterbodies, do nothing to the edge and move on to the next edge in the list.
      \item If the immediate upstream nodes are not waterbodies, test if all upstream nodes and the TOCOMID node are in the same HUC12 watershed. If they are, then replace all instances of the FROMCOMID river node in the list of TOCOMIDs and FROMCOMIDs with the TOCOMID node. If they are not in the same HUC12 watershed, do nothing to the edge and move on to the next edge in the list.
    \end{itemize}
    \item If there is not another immediate upstream node to the TOCOMID node, then test if the FROMCOMID and TOCOMID node are in the same HUC12 watershed.
    \begin{itemize}
      \item If they are in the same HUC12 watershed, replace all instances of the FROMCOMID river node in the list of TOCOMIDs and FROMCOMIDs with the TOCOMID node. 
      \item If they are not in the same HUC12 watershed, do nothing to the edge and move on to the next edge in the list. 
    \end{itemize}
  \end{itemize}
\end{itemize}

\begin{figure}[!htp]
  \centering
  \includegraphics[scale=.6]{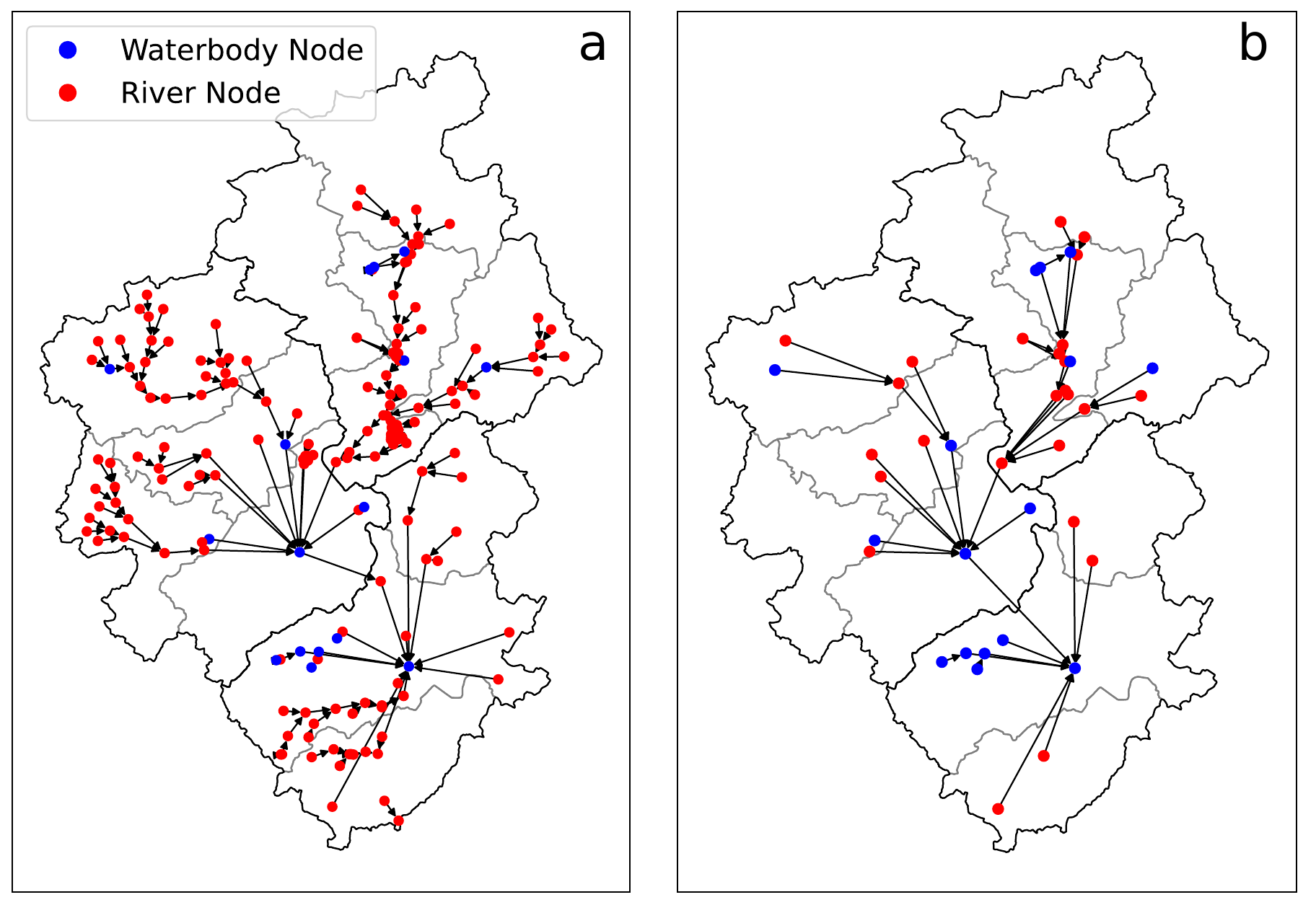}
  \caption{A visualization of the aggregation of our graph. Nodes of the original graph (panel a) are iteratively merged to form the aggregated graph (panel b). Watersheds shown are the the HUC10 watersheds 0709000205, 0709000206, and 0709000207, primarily in Dane County, Wisconsin.}
  \label{fig:aggregation}
\end{figure}

This algorithm is repeated continuously until it ceases to change the set of TOCOMIDs and FROMCOMIDs. In the case of the graph of Wisconsin, this method reduced our graph from 45,997 nodes to 8,526 nodes. The results of this aggregation procedure can be seen in Figure \ref{fig:aggregation}.\\

We make two notes on this algorithm. First, in several steps of the algorithm, we are concerned with the immediate upstream or downstream connections to a given node. This is because we have to be careful not to merge these connections as they would change the connectivity of the graph. For example, imagine a waterbody with an upstream node that flows both into the waterbody and into a separate river node. If we merge that upstream node into the waterbody, the waterbody would now also flow into the separate river node. This would not maintain the overall connectivity of the original graph. Second, this algorithm also highlights one of the reasons that we added the HUC watershed codes as attributes of the waterbodies shapefile since we use the HUC12 codes to only merge nodes in the same watershed. These attributes (such as the watershed codes) can be important in working with these objects. 
\\

In aggregating our graph, we wanted to confirm that the connectivity of the original graph was maintained. In particular, it is important that if two waterbodies are connected in the original graph, they should also be connected in the aggregated graph. To confirm this was the case, we looped through every waterbody within the aggregated graph in an outer loop, and then looped through the same set of waterbodies within an inner loop. Inside the inner loop, we used the function {\tt network.has\_path} to test if the waterbody of the outer loop and the waterbody of the inner loop are connected within a given graph. We ran this test for both the original and aggregated graph. After running this function, we tested whether the original graph and the aggregated graph yielded the same results. This test can be found at the end of "Visualizations.ipynb". It looped over 3,199 waterbodies, and the connectivity was the same between the original and aggregated graphs for all waterbodies.

\section{Case Study I: Lake Altoona and Mohawksin Lake Data}

In this section, we report the data used for Lake Mohawksin and Altoona Lake as discussed in case study 1 of our manuscript. We report the individual measurements for total phosphorus (TP) and the lake perception (Tables \ref{tab:Table1} - \ref{tab:Table4}) taken from the Wisconsin Department of Natural Resources \cite{dnraltoona,dnrmohawksin}. This latter measurement is a 1-5 scale of algae levels on the lake, with 1 ("beautiful, could not be nicer") representing cleaner levels and 5 ("swimming and aesthetic enjoyment of lake substantially reduced because of algae levels") representing poorer water quality. The lake perception can give a general idea of water quality levels, but it is subjective (based on the opinion of the data collector) and has no quantitive measure \cite{dnr_lake_monitoring}. Overall, Altoona Lake had much higher TP measurements, and there were also higher perception levels reported. Lake Mohawksin never had a perception level above 2 while Altoona Lake had several reported levesl of 3 and 4. The data strongly suggests that Altoona Lake has poorer water quality than Lake Mohawksin. 

\begin{table}
\caption{Total phosphorus measurements for Altoona Lake, Wisconsin}
\begin{center}
\begin{filecontents*}{altoona_lake_tp.csv}
Date,TP (mg/m3),,Date,TP (mg/m3)
5/15/2001,79,,6/27/2015,85.9
7/31/2001,75,,7/24/2015,95.5
8/8/2011,142,,8/29/2015,87.9
8/29/2011,72,,5/9/2016,64.1
5/24/2013,127,,6/25/2016,129
7/6/2013,138,,7/23/2016,52.2
7/29/2013,80.3,,9/6/2016,160
8/24/2013,92.7,,6/26/2017,142
5/10/2014,71.7,,5/7/2018,113
6/27/2014,123,,6/28/2018,123
7/26/2014,80.2,,7/29/2018,117
8/30/2014,94.6,,8/23/2018,166
4/18/2015,67.4,,6/22/2019,98.3
\end{filecontents*}
\csvautotabular{altoona_lake_tp.csv}
\end{center}
\label{tab:Table1}
\end{table}

\begin{landscape}
\begin{table}
\begin{center}
  \caption{Perception measurements for Altoona Lake, Wisconsin}
  \begin{filecontents*}{altoona_lake_perception.csv}
Date,Perception,,Date,Perception
5/12/2011,3-Enjoyment somewhat impaired (algae),,7/16/2019,2-Very minor aesthetic problems
8/29/2011,4-Would not swim but boating OK (algae),,7/24/2019,2-Very minor aesthetic problems
9/15/2011,4-Would not swim but boating OK (algae),,8/1/2019,2-Very minor aesthetic problems
5/19/2018,2-Very minor aesthetic problems,,8/27/2019,1-Beautiful
6/12/2018,3-Enjoyment somewhat impaired (algae),,4/11/2020,1-Beautiful
6/20/2018,2-Very minor aesthetic problems,,4/27/2020,1-Beautiful
6/28/2018,2-Very minor aesthetic problems,,5/24/2020,1-Beautiful
7/14/2018,2-Very minor aesthetic problems,,5/31/2020,1-Beautiful
7/22/2018,4-Would not swim but boating OK (algae),,6/6/2020,1-Beautiful
7/29/2018,4-Would not swim but boating OK (algae),,6/8/2020,1-Beautiful
8/14/2018,4-Would not swim but boating OK (algae),,6/17/2020,1-Beautiful
8/23/2018,4-Would not swim but boating OK (algae),,6/24/2020,2-Very minor aesthetic problems
8/31/2018,4-Would not swim but boating OK (algae),,7/5/2020,2-Very minor aesthetic problems
9/8/2018,2-Very minor aesthetic problems,,7/11/2020,2-Very minor aesthetic problems
9/16/2018,2-Very minor aesthetic problems,,7/19/2020,2-Very minor aesthetic problems
9/23/2018,2-Very minor aesthetic problems,,7/27/2020,2-Very minor aesthetic problems
4/14/2019,1-Beautiful,,8/9/2020,3-Enjoyment somewhat impaired (algae)
4/23/2019,2-Very minor aesthetic problems,,8/11/2020,3-Enjoyment somewhat impaired (algae)
5/4/2019,1-Beautiful,,8/20/2020,4-Would not swim but boating OK (algae)
5/16/2019,2-Very minor aesthetic problems,,8/27/2020,4-Would not swim but boating OK (algae)
6/6/2019,2-Very minor aesthetic problems,,9/13/2020,2-Very minor aesthetic problems
6/14/2019,2-Very minor aesthetic problems,,9/21/2020,2-Very minor aesthetic problems
6/22/2019,2-Very minor aesthetic problems,,9/30/2020,2-Very minor aesthetic problems
7/2/2019,2-Very minor aesthetic problems,,10/7/2020,2-Very minor aesthetic problems
7/8/2019,2-Very minor aesthetic problems,,,
\end{filecontents*}
\csvautotabular{altoona_lake_perception.csv}
  \end{center}
\end{table}
\end{landscape}

\begin{table}
\begin{center}
  \caption{Total phosphorus measurements for Lake Mohawksin, Wisconsin}
  \begin{filecontents*}{mohawksin_lake_tp.csv}
Date,TP (mg/m3),,Date,TP (mg/m3)
7/17/2006,37,,9/4/2013,38.9
8/15/2006,48,,5/17/2014,36.5
10/15/2006,36,,6/26/2014,34.7
5/1/2007,32,,7/14/2014,38.3
6/29/2007,38,,7/24/2014,45
7/16/2007,42,,8/28/2014,33.4
8/23/2007,38,,6/24/2015,37.5
5/8/2008,34,,8/26/2015,39.1
6/25/2008,35,,5/22/2016,32.4
8/6/2008,38,,6/22/2016,43.1
8/29/2008,35,,7/25/2016,49.6
5/2/2009,39,,9/2/2016,40.2
6/20/2009,30,,6/26/2017,52.3
8/3/2009,39,,7/19/2017,53.4
9/8/2009,20,,8/21/2017,42.3
5/25/2010,25,,5/27/2018,26.7
6/30/2010,48,,6/22/2018,45.9
8/10/2010,75,,7/23/2018,48.3
6/29/2011,46,,8/16/2018,34
7/29/2011,53,,4/29/2019,29.6
9/14/2011,46,,5/23/2019,31.3
5/1/2012,42,,6/24/2019,42
6/27/2012,34,,7/29/2019,39.6
7/31/2012,48,,8/19/2019,36.9
8/29/2012,40,,6/25/2020,32.8
5/12/2013,43.1,,7/27/2020,44.7
6/25/2013,32.4,,8/31/2020,46.7
7/29/2013,44.6,,,
\end{filecontents*}
\csvautotabular{mohawksin_lake_tp.csv}
    \end{center}
  \end{table}
  
\begin{table}
\begin{center}
    \caption{Perception measurements for Lake Mohawksin, Wisconsin}
    \begin{filecontents*}{mohawksin_lake_perception.csv}
Date,Perception,,Date,Perception
7/17/2006,2-Very minor aesthetic problems,,5/17/2014,1-Beautiful
7/31/2006,2-Very minor aesthetic problems,,6/26/2014,1-Beautiful
8/15/2006,2-Very minor aesthetic problems,,7/24/2014,2-Very minor aesthetic problems
9/4/2006,2-Very minor aesthetic problems,,8/28/2014,1-Beautiful
9/20/2006,2-Very minor aesthetic problems,,6/24/2015,2-Very minor aesthetic problems
10/2/2006,2-Very minor aesthetic problems,,7/14/2015,2-Very minor aesthetic problems
10/15/2006,2-Very minor aesthetic problems,,8/26/2015,2-Very minor aesthetic problems
5/8/2008,2-Very minor aesthetic problems,,5/22/2016,2-Very minor aesthetic problems
5/28/2008,2-Very minor aesthetic problems,,6/22/2016,2-Very minor aesthetic problems
6/25/2008,2-Very minor aesthetic problems,,9/2/2016,2-Very minor aesthetic problems
7/14/2008,2-Very minor aesthetic problems,,6/26/2017,2-Very minor aesthetic problems
9/17/2008,2-Very minor aesthetic problems,,7/19/2017,2-Very minor aesthetic problems
5/3/2009,2-Very minor aesthetic problems,,8/21/2017,2-Very minor aesthetic problems
6/20/2009,2-Very minor aesthetic problems,,5/27/2018,2-Very minor aesthetic problems
6/5/2011,1-Beautiful,,5/27/2018,2-Very minor aesthetic problems
6/29/2011,2-Very minor aesthetic problems,,6/22/2018,2-Very minor aesthetic problems
7/29/2011,2-Very minor aesthetic problems,,7/23/2018,2-Very minor aesthetic problems
9/14/2011,2-Very minor aesthetic problems,,8/16/2018,2-Very minor aesthetic problems
5/1/2012,2-Very minor aesthetic problems,,5/23/2019,2-Very minor aesthetic problems
6/27/2012,2-Very minor aesthetic problems,,6/24/2019,2-Very minor aesthetic problems
7/31/2012,2-Very minor aesthetic problems,,7/29/2019,2-Very minor aesthetic problems
8/29/2012,2-Very minor aesthetic problems,,8/19/2019,2-Very minor aesthetic problems
5/12/2013,2-Very minor aesthetic problems,,6/6/2020,2-Very minor aesthetic problems
6/25/2013,2-Very minor aesthetic problems,,7/7/2020,2-Very minor aesthetic problems
7/29/2013,2-Very minor aesthetic problems,,8/31/2020,2-Very minor aesthetic problems
9/4/2013,2-Very minor aesthetic problems,,,
\end{filecontents*}
\csvautotabular{mohawksin_lake_perception.csv}
    \label{tab:Table4}
      \end{center}
\end{table}

\section{Case Study II: DNR Data}

\subsection{Data Compilation}

The data used in Case Study 2 of our manuscript was taken from the Wisconsin Department of National Resources's (DNR) website \cite{dnr} on 10 August 2021. We wrote a script to webscrape and download the data of interest. Many of the measurements reported by the DNR were taken at different points within a given lake. For consistency, we only compiled data where the measurements were taken near the deepest point of the lake (i.e., only where the title of the data included the words "Deep Hole", "Max Depth", "Deepest", or "Maximum Depth"). The resulting reports were reformatted so that the TP and chlorophyll-a data was more accessible. The scripts for completing this process are availabe at \url{https://github.com/zavalab/ML/tree/master/HydroGraphs/DNR_data}. The original lake reports that we downloaded ("original\_lake\_reports/") and the cleaned, reformatted reports ("Lakes/") are also available at this link. This folder ("Lakes/") is indexed by "lake\_index\_WBIC\_COMID.csv". \\

In addition, one of the challenges of using this data is matching the data to the correct waterbody within the NHDPlusV2. The DNR data uses a Waterbody ID (WBIC) while the NHDPlusV2 uses a COMID. To match these, we needed a set of shapefiles that identified the lakes by the WBIC. We use the Wisconsin DNR's 24k Hydro Waterbodies dataset \cite{DNRhydrowaterbody} dataset which uses a WBIC to identify all of the waterbody polygons. To match the NHDPlusV2 with the 24k Hydro Waterbodies, we first took all 24k Hydro Waterbodies which had a WBIC that belonged to any of the DNR data which we compiled. This resulted in a set of about 800 waterbodies. We then iterated through this set of waterbodies and tested whether the centroid of any of these HYDROLake polygons was within a NHDPlusV2 waterbody. If they were, we saved the COMID of that NHDPlusV2 waterbody so that it corresponded to the WBIC of the HYDROLake. Because of the shape of some of the waterbodies, the some centroids lie outside of the given waterbody. Consequently, for all of those waterbodies for which the centroid did not lie within a NHDPlusV2 waterbody, we tested whether the HYDROLake polygon intersected any NHDPlusV2 waterbody. If it only intersected a single NHDPlusV2 waterbody, we also saved the COMID of the NHDPlusV2 waterbody so that it corresponded to the WBIC of the intersecting HYDROLake. The code for performing this task can be found in the above github link, along with the HYDROLakes shapefiles that we used. 

\subsection{Cutoff Values for Analysis}

In analyzing the waterbodies with the highest and lowest pollutant concentrations, we used cutoff values to help ensure that both the TP and chlorophyll-a data suggested that the waterbodies were polluted. Carlson \cite{Carlson1977trophic} introduced a Trophic State Index (TSI) for helping to identify the eutrophication level (strongly related to the water quality) of a waterbody. The TSI of a waterbody can be calculated from the secchi depth, total phosphorus level, or chlorophyll-a level. Carlson also points out that the calculated TSI value should be the same regardless of what initial indicator is used. Consequently, we chose the cutoffs to ensure that both values would be above a certain threshold. In addition, Carlson and Simpson \cite{carlson1996coordinator} suggest general ranges for classifying a lake as oligotrophic or eutrophic (essentially good or poor water quality, respectively). 

We calculated the TSI for the cutoffs used in this study, and the resulting values suggest that the chosen cutoffs are reasonable. For identifying the five most and least polluted levels, the cutoffs for the least polluted lakes require that the TSI be below 47 in both TP and chlorophyll-a (based on Carlson and Simpson \cite{carlson1996coordinator}, this is below the classical eutrophy range). For the most polluted lakes, the cutoffs require the TSI to be above 63. However, we note that for the lakes chosen, all of the least polluted lakes had a TSI---based on average TP and chlorophyll-a values---below 40. The most polluted lakes in contrast had a TSI of no lower than 65 (well into the eutrophic range reported by Carlson and Simpson), and most lakes had a TSI of over 70 (likely making them hypereutrophic). For the cutoffs used to form a larger subset of lakes, the values of $<$ 15 mg/m\textsuperscript{3} TP and $<$ 5 mg/m\textsuperscript{3} chlorophyll-a correspond to a TSI no larger than 47 for the less polluted lakes. For the more polluted lakes, the cutoffs of $>$ 60 mg/m\textsuperscript{3} TP and $>$ 15 mg/m\textsuperscript{3} chlorophyll-a correspond to a TSI of no smaller than 57.\\

\section{Urban Land Fraction Calculations}
Curating the data for urban land was different than that done for the agricultural land. All datasets we found for urban land cover were in a raster format (rather than a shapefile format) making them more difficult for us to work with in {\tt GeoPandas}. We considered using population data combined with election data wards, but population alone may not perfectly represent areas of high urban landcover (e.g., industrial sectors may not match areas of dense population). We chose to use the Wiscland2 dataset \cite{dnrwiscland2} from the Wisconsin DNR which classifies all land in Wisconsin, including an "urban land" classification. We converted this data from a raster to a shapefile, and a script for performing this conversion are available in our repository. The documentation for the Wiscland2 dataset reports that the coordinate system is the Wisconsin Transverse Mercator NAD83 (HARN) which, as we understand it, corresponds to EPSG code 3071. When we convert from EPSG:3071 to EPSG:4326 (which is the projection in which we perform the majority of our analysis), the resulting shapefile contains an artifact (we could visually see that the projected data had a band down the middle) which was not visibly present under the original projected data. Consequently, when we performed any computations on urban land cover, we maintained the wiscland2 shapefile in the Wisconsin Transverse Mercator NAD83 (HARN) Projection and converted the other shapefiles to this projection for performing urban land fraction calculations to avoid any miscalculations caused by the artifact. Thus, in the Case Study 1 and Case Study 2 scripts, the urban land fraction computations are all performed under the EPSG:3071 code (with accompanying shapefiles converted to that projection) while agricultural land fraction computations were performed under the EPSG:4326 code as the agricultural land came from a different dataset without an artifact. 

\section{Additional Functions for working with Graphs}\label{sec:functions}

In this section, we give some of the functions that we have defined in "HydroGraph\_functions.py" at \url{https://github.com/zavalab/ML/tree/master/HydroGraphs/graph_construction} to make managing and building the graphs easier. See the {\tt NetworkX} documentation for further details on how to use specific functions within their package.

\begin{itemize}
  \item {\tt build\_graph(tofroms)}--This function takes the dataframe that contains the list of TOCOMID and FROMCOMID values for a given graph and converts it into a directed graph within {\tt NetworkX}. It does this by adding edges to the graph from the TOCOMID and FROMCOMID values. 
  \item {\tt get\_pos\_dict(G, lake\_gdf, river\_gdf)}--This function takes a graph and the GeoDataFrames for all waterbodies and rivers within the graph and returns a dictionary of positions (based on the centroid of the objects in the GeoDataFrames), a list of node colors (with a default of blue for waterbodies and red for rivers), and a list of node sizes. This information is useful in plotting within {\tt NetworkX}'s {\tt draw()} function. 
  \item {\tt get\_upstream\_graph(G, node)}--This function returns the upstream graph for a specified node.
  \item {\tt get\_downstream\_graph(G, node)}--This function returns the downstream graph for a specified node.
  \item {\tt get\_upstream\_graph\_and\_cols(G, node, lake\_gdf, river\_gdf)}--This function returns the upstream graph for a specified node, and it returns a list of colors that can be used for plotting the upstream graph.
  \item {\tt get\_downstream\_graph\_and\_cols(G, node, lake\_gdf, river\_gdf)}--This function returns the downstream graph for a specified node, and it returns a list of colors that can be used for plotting the downstream graph. 
\end{itemize}

For specific details of how these functions operate, please see the comments within the file containing these functions.

\bibliography{supplementary_material}